\documentclass[final,1p,times,twocolumn]{elsarticle}

\usepackage{amssymb}
\usepackage{amsmath}

\usepackage{url}
\usepackage{amsmath,amssymb,amsfonts}
\usepackage{array}
\usepackage{algorithmic}
\usepackage{stfloats}
\usepackage[caption=false,font=normalsize,labelfont=sf,textfont=sf]{subfig}
\usepackage{makecell}
\usepackage{framed,multirow}

\journal{EXPERT SYSTEMS WITH APPLICATIONS}

\begin{document}

\begin{frontmatter}



\title{Adaptive Critical Subgraph Mining for Cognitive Impairment Conversion Prediction with T1-MRI-based Brain Network}

\author[1,2]{Yilin Leng}
\author[1,2]{Wenju Cui}
\author[1,2]{Bai Chen}
\author[3]{Xi Jiang}
\author[4]{Yunsong Peng\corref{cor1}}
\ead{pys@mail.ustc.edu.cn}
\author[1,2,5]{Jian Zheng\corref{cor1}}
\ead{zhengj@sibet.ac.cn}
\cortext[cor1]{Corresponding author}

\address[1]{School of Biomedical Engineering (Suzhou), Division of Life Sciences and Medicine, University of Science and Technology of China, Hefei 230026, China}
\address[2]{Department of Medical Imaging, Suzhou Institute of Biomedical Engineering and Technology, Chinese Academy of Sciences, Suzhou 215163, China}
\address[3]{Clinical Hospital of Chengdu Brain Science Institute, MOE Key Lab for Neuroinformation, School of Life Science and Technology, University of Electronic Science and Technology of China, Chengdu 611731, China}
\address[4]{Medical imaging department, Guizhou Provincial People's Hospital, Guizhou 550002, China}
\address[5]{Shandong Laboratory of Advanced Biomaterials and Medical Devices in Weihai, Weihai 264200, China}



\begin{abstract}
Prediction conversion of early-stage dementia is challenging due to subtle cognitive and structural brain changes.
Traditional T1-weighted magnetic resonance imaging (T1-MRI) studies primarily identify brain atrophy but often overlook the intricate inter-regional connectivity, limiting a comprehensive understanding of the brain's complex network.
Moreover, there’s a pressing demand for methods that can adaptively preserve and extract critical information, particularly specialized subgraph mining techniques for brain networks.
Such methods are essential for developing high-quality feature representations that reveal spatial impacts of structural brain changes and their topology.
In this paper, we propose Brain-SubGNN, a novel graph representation network to mine and enhance critical subgraphs on T1-MRI. This network provides a subgraph-level interpretations to improve interpretability and graph analysis insights.
The process begins by extracting node features and a correlation matrix between nodes to construct a task-oriented brain network.
Brain-SubGNN then adaptively identifies and enhances critical subgraphs, capturing both loop and neighbor subgraphs.
This method reflects the loop topology and local changes, indicative of long-range connections, and maintains local and global brain attributes.
Extensive experiments validate the effectiveness and advantages of Brain-SubGNN, demonstrating its potential as a powerful tool for understanding and diagnosing early-stage dementia. 
Source code is available at \url{https://github.com/Leng-10/Brain-SubGNN}.
\end{abstract}


\end{frontmatter}


\section{Introduction}
\label{sec:introduction}
\begin{figure}[!ht]
\centering
\includegraphics[width=4in]{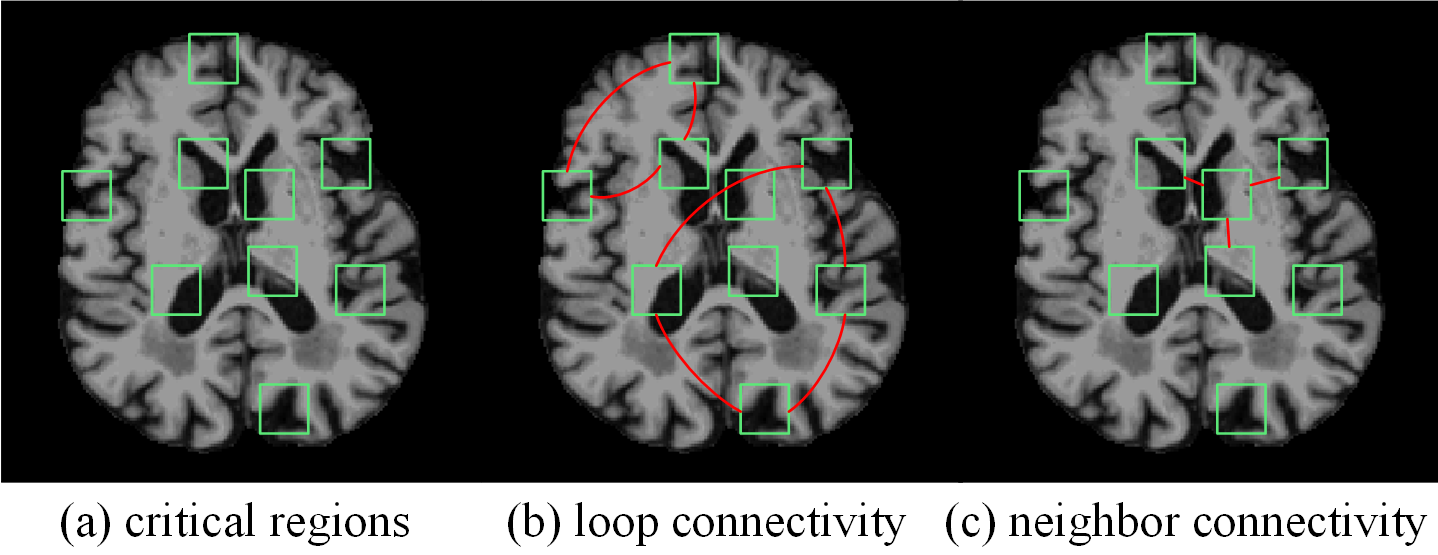}%
\caption{
There are some changes in brain through cognitive impairment.
(a) critical regions related to atrophy or some other lesion, 
(b) closed-loop mechanisms in brain self-regulation, 
(c) local neighbor connectivity. }
\label{fig_progreesion}
\end{figure}

Progressive cognitive impairment is considered a severe and significant subtype of dementia, including progressive mild cognitive impairment (pMCI) and the less explored progressive normal controls (pNC) \cite{metastasio2006conversion, yue2020characterizing}. 
These symptoms represent a continuum of cognitive decline, ranging from subtle changes to more severe impairments. 
T1-weighted Magnetic resonance imaging (T1-MRI) has emerged as a crucial non-invasive, radiation-free modality for early dementia assessment\cite{frisoni2010clinical}.
It offers superior soft tissue contrast, ideal for identifying and visualizing brain atrophy  \cite{chincarini2011local}.

Recent deep learning (DL) methods utilizing T1-MRI \cite{liu2018landmark, lian2020attention, pan2021disease} have advanced the assessment of brain atrophy, holding potential for intelligent assessment of cognitive impairment. 
However, despite the state-of-the-art (SOTA) performance of Convolutional Neural Networks (CNNs),
current TI-MRI-based methods often compromise the global brain shape and topology \cite{lian2018hierarchical, guan2021multi, chen2022alzheimer, han2023multi}.
This is due to methods that reduce irrelevant context by cropping equal-sized patches \cite{lian2018hierarchical} or focusing solely on discriminative anatomical brain regions, such as hippocampus \cite{zhang2023multi, zhu2022interpretable}. 
To address these challenges, several studies\cite{bullmore2009complex, thiebaut2022emergent, axer2022scale, zhu2022interpretable} have constructed brain networks incorporating shaped-related priors.
However, these methods detect fuzzy patterns and struggle to capture precise distinguishable information effectively.
Therefore, localizing critical structures (e.g., subgraphs) is essential for analyzing and interpreting changes in the brain.

Preserving and extracting maximum information is essential for high-quality feature representations, particularly in pinpointing the spatial distribution of effects caused by local structural changes. 
The advancement of Graph Neural Networks (GNNs) has broadened applications across various fields, including human networks \cite{alvarez2021evolutionary}, traffic analysis \cite{chen2024traffic}, and drug discovery \cite{fang2022geometry}.
While numerous studies \cite{song2021graph, song2022multi, lei2020self, chen2022adversarial, li2022joint} have successfully employed GNNs for analyzing human brain networks, 
most \cite{duran2022dual, leng2023dynamic, cui2021brainnnexplainer, kan2022fbnetgen, zhu2022joint, cui2022interpretable} focus on graph or node classification tasks, overlooking critical subgraphs.
Existing subgraph mining methods, like graph kernels \cite{kriege2020survey, shervashidze2011weisfeiler}, decompose pre-defined structures from the original graph, demonstrating competitive performance in specific domains.
However, their handcrafted and heuristic nature limits their flexibility and generalization capabilities. 
Meanwhile, recent graph transformers \cite{chen2022structure, geisler2023transformers} struggle to identify structural similarities between nodes, limiting their ability to identify critical edges.
Therefore, the quest to adaptively identify critical structures at both local and global scales, without fixed rules or prior knowledge, remains a challenge.

Recent studies \cite{sitaram2017closed, skouras2020earliest} highlight the roles of specific brain regions in neurofeedback control and reward processing, emphasizing the importance of underlying closed-loop mechanisms in brain self-regulation. 
These insights inform the critical need for loop information mining in neurological research.
Moreover, the inherent rich club \cite{van2011rich} and small-world \cite{bassett2006small} properties of brain networks, underlining local and long-range connectivity, are underexplored in T1-MRI-based research.
These properties are crucial for understanding the complex interplay of brain regions in cognitive disorders. 
Consequently, there is a significant gap in the development of subgraph mining methods that can adaptively explore data-specific and task-specific critical structures in brain networks, i.e., subgraphs of arbitrary size and shape.

\begin{figure*}[!t]
\centering
\includegraphics[width=5.4in]{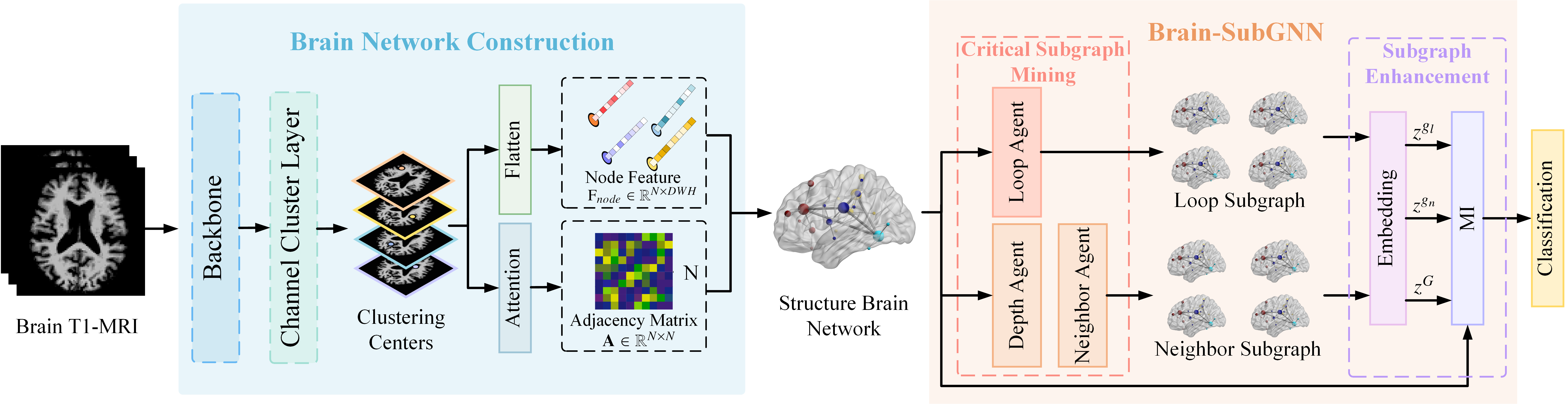}%
\caption{The workflow of our proposed method. It assesses brain disorders in three steps:
1) constructing adaptively data-specific and task-specific structural brain networks with both dynamic nodes and dynamic connections;
2) mining critical subgraphs through two Modules, which adaptively mine local loop and neighbor subgraphs of task-oriented, individually heterogeneous and arbitrarily size and shape; 
3) encoding loop and neighbor subgraphs, and fuses subgraphs information with global graphs using shallow graph convolution for downstream tasks.}
\label{fig_archi}
\end{figure*}

In this paper, we present a novel approach called Brain-SubGNN for early diagnosis of cognitive impairment, addressing these challenges with a brain subgraph neural network. 
The pipeline of our framework is illustrated in Fig.~\ref{fig_archi}, aims to enhance the identification of critical structures and improve interpretability.
Firstly, we adopt a unified graph representation that adaptively allocates nodes to discriminative regions, thereby accounting for varying atrophic patterns and connectomes across individuals. 
Secondly, we design a subgraph mining network named Brain-SubGNN to acquire graph topology information and mine critical subgraphs without prior knowledge. 
We innovatively incorporate reinforcement learning (RL) based subgraph mining modules for dynamic identification of critical neighbor and loop subgraphs associated with cognitive impairment. 
Additionally, we compute mutual information (MI) between all mined subgraphs and the global graph to reinforce subgraphs representation.

The experimental results demonstrate that our novel approach outperforms several SOTA methods in predicting the conversion of MCI (i.e., sMCI vs. pMCI). Notably, we make a pioneering contribution by successfully predicting the likelihood of future conversion from Normal Controls to MCI (sNC vs. pNC), presenting promising results.  
Our contributions are illustrated as followings:
\begin{itemize}
\item This work constructs a complete multi-resolution framework, providing node-, subgraph-, and graph-level representation.
\item Brain-SubGNN is a data and task-driven subgraph mining network. 
A loop and a neighbor subgraph mining module are designed to detect changes in self-regulation and local effects associated with cognitive degradation.
\item This study reports a remarkable achievement in identifying individuals who will develop cognitive impairment within three years from NC on publicly available datasets. 
This suggests the potential to further advance early diagnosis of cognitive impairment.
\end{itemize}

\section{Related Works}
\label{sec:relatedworks}

\subsection{Methods on T1-MRI and Brain Networks}
Recently, DL methods have demonstrated remarkable performance in computer vision tasks, prompting their exploration in computer-aided diagnosis of neurodegenerative diseases. Four primary types of DL methods are illustrated as: 
\textit{1) Whole-image-based methods}: a widely employed class of DL techniques, extract spatial contextual information directly from original images \cite{pan2021disease}; 
\textit{2) ROI-based methods}: partition brain imaging data into regions of interest (ROI) using templates like automatically Automated Anatomical Labeling (AAL) \cite{ashburner2000voxel} or manual anatomy for analysis and classification \cite{zhang2011multimodal};
\textit{3) Slice-based methods}: divide the image into slices and employ a 2D network for classification \cite{chen2022alzheimer};
\textit{4) Patch-based methods}: crop patches or use attention mechanisms to focus on regions affected by early dementia in the MR images \cite{liu2018landmark, lian2020attention, guan2021multi}.

The brain's complexity and diversity make graphical representation ideal for analyzing MR images' intricate backgrounds and heterogeneous lesions.
While CNNs are proficient with Euclidean data, they struggle with non-Euclidean data like the brain's macrostructural complexity \cite{bronstein2017geometric}. 
Graph-based methods transform brain MR images into graphs for region identification and connection analysis overcome the challenge associated with applying fundamental mathematical procedures.
In the realm of fMRI-based research \cite{chen2022adversarial, song2022multi, lei2020self, li2022joint, duran2022dual, cui2022interpretable}, such methods have been widely employed to construct brain function connection network for the analysis of disease processing, consistently achieving SOTA performance cross multiple datasets.
However, applying GNNs to T1-MRI is still in development due to the absence of direct correlation between regions.

\subsection{Brain Subgraph Mining}
The success of GNNs in analyzing graph-structured data \cite{alvarez2021evolutionary, chen2024traffic, fang2022geometry} has led to deep models predicting brain diseases by learning graph structures of brain networks \cite{cui2021brainnnexplainer, kan2022fbnetgen, zhu2022joint, cui2022interpretable}. 
Conventional approaches, like graph kernels (e.g., temporal random walk \cite{gartner2003graph, wang2021inductive}, shortest path \cite{borgwardt2005shortest}, neighborhood representation \cite{luo2022neighborhood} and subgraph sketching), decompose pre-defined structures from the original graph, demonstrating competitive performance in specific domains.
However, their handcrafted and heuristic nature limits their flexibility and generalization capabilities. 
Meanwhile, recent graph transformers \cite{cong2023we} struggle to identify structural similarities between nodes, limiting their ability to identify critical edges.
Therefore, detecting data-specific and task-specific critical structures at both local and global scales in brain network without fixed rules or prior knowledge remains an unsolved problem.

Recently, RL has been incorporated into various graph mining tasks \cite{tang2023systematic}, such as graph representation learning \cite{lai2020policy}, graph adversarial attack \cite{xu2020adversarial}, relational reasoning \cite{zhu2019causal}, GNN explainer \cite{yuan2021explainability}, and combination optimization \cite{mazyavkina2021reinforcement}.
RL is a fundamental branch of machine learning where agents learn to make optimal decision sequentially within an environment. 
The mathematical framework of Markov decision process (MDP) is commonly employed to formulate RL as a sequential decision process \cite{wang2020second}.
Similar to Brain-SubGNN, some methods \cite{dou2020enhancing, peng2022reinforced} use RL to identify critical structures and enhance graph learning. 
For instance, Lai et al. \cite{lai2020policy} employed the Deep Q-Network (DQN) \cite{mnih2015human} to determine the optimal number of aggregation iterations for different nodes, 
while Gao et al. \cite{gao2019graphnas} used RL to search for optimal GNN architectures, showing RL's growing application in GNN optimization and development.

In conclusion, existing GNNs with RL focus on identifying critical relations or improving GNN architectures for node-level tasks. However, limited research has been conducted to advanced RL for identifying more complex critical structures in graph-level tasks.

\section{Methods}

\subsection{Dynamic Brain Network Construction}
\label{sec:construct}

To construct brain networks from T1-MRI data, we utilize methods from our prior work \cite{leng2023dynamic} to identify discriminative regions and examine their connectivities. 
Each subject is conceptualized as a graph $G=\{V,A\}$, 
where $V$ represents vertices corresponding to discriminative regions, and $A$ denotes the edges reflecting the relationships between these regions. 
This results in a brain network adaptively tailored to the specific characteristics of each subject.

\subsubsection{\textbf{Selection of Nodes}}
We first obtain abundant feature maps using a powerful encoder, then cluster the feature maps of multiple channels to obtain discriminative feature maps as nodes.
Specifically, peak responses coordinates of feature maps are defined as the candidate prototypes inspired by \cite{zheng2017learning}, then feature maps with representative positions are selected as discriminative ones by K-means \cite{lloyd1982least}.
The position vectors of feature maps over all training images are expressed as:
\begin{equation}
[t_x^1,t_y^1,t_z^1,\dots,t_x^{\Omega},t_y^{\Omega},t_z^{\Omega}]
\end{equation}
where $[t_x^i,t_y^i,t_z^i]$ denotes the peak response coordinate of the feature map corresponding to the $i$-th images, and $\Omega$ is the number of training images. Clustering these peak responses using K-means yields $N$ clustering centers, serving as nodes $V=\{v_1,v_2,\dots,v_N\}$ for graph construction, where $v_j$ is the feature map of the $j$-th channel clustering center. 

To integrate clustering process into the network training and optimize the results, we employ a channel clustering layer (CCL) with prototype contrastive learning into network training. 
Specifically, the CCL comprises two fully connected layers (FCs) and a skip connection, encouraging each channel cluster to be compact internally and have significant inter-class differences from other clusters. 
Optimization is performed using the prototype contrastive learning \cite{li2020prototypical} loss function:
\begin{equation}
\mathcal{L}_{ccl}=-\frac{1}{N_c}\sum_{i=1}^{N_c}\sum_{u\in{K_i}}\log\frac{exp(u\cdot\gamma_i/\phi_i)}{\sum_{j\neq i}^{N}exp(u\cdot\gamma_j/\phi_i)}
\label{L_ccl}
\end{equation}
\begin{equation}
\phi_i=\frac{\sum_{u\in{K_i}}{\|u-\gamma_i\|}_2}{|K_i|\cdot \log(|K_i|+\alpha)}
\label{phi}
\end{equation}
Where $N_c$ is the number of clusters, $K_i$, $\gamma_i$, and $\phi_i$ denote the set of all elements, the cluster center (prototype), and the concentration estimation of the $i$-th cluster, respectively. $\alpha$ is a smoothing parameter to prevent small clusters from having overly-large $\phi$. The cluster concentration $\phi$ measures the closeness of elements in a cluster. 
$\mathcal{L}_{ccl}$ compels all elements $u$ in $K_i$ to be close to their cluster center $\gamma_i$ and away from other cluster centers. 

In this way, a data- and task-oriented discriminative regions are automatically identified as nodes for each subject.

\subsubsection{\textbf{Generation of Edges}}
To eatablish adaptive inter-regional correlations, we employ an attention mechanism \cite{vaswani2017attention} for edge connection in each image, allowing nodes to exchange features along related graph edges. 
Specifically, the clustering feature $F_{ccl}\in \mathbf{R}^{N\times D\times W\times H}$ obtained from CCL are processed through three sperate FCs to compute attention scores between each pair of prototypes. The edge connection is stored by the adjacency matrix $A\in \mathbf{R}^{N\times N}$ to obtain the output of self-attention layer:
\begin{equation}
A=Attention(Q,K,V)=softmax(\frac{QK^T}{\sqrt{d_k}})V
\end{equation}
where $Q\in \mathbf{R}^{N\times d_k}$, $K\in \mathbf{R}^{N\times d_k}$ and $V\in \mathbf{R}^{N\times N}$ denote query, key, and value, respectively. $d_k$ represents the dimension of $Q$ and $K$. $N$ is the number of discriminative regions.

Ultimately, by employing the prototypes as nodes and correlations as edges, we construct a disease-related individualized structural brain network with both adaptive nodes and adaptive connections.

\begin{figure*}[!ht]
\centering
\includegraphics[width=5.5in]{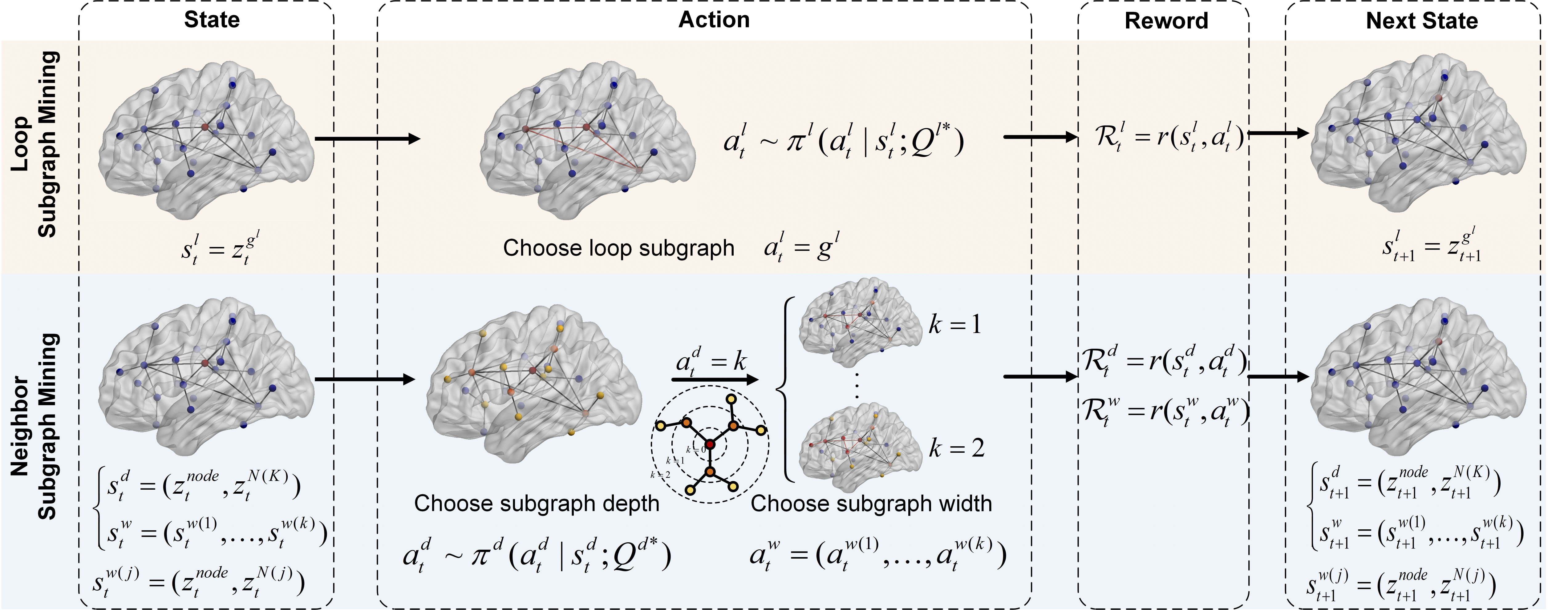}%
\caption{The workflow of subgraph mining module. For a given center node, 1) the LSRM generates an action $a_t^l$ by policy $\pi^l$ to detect the critical loop subgraph; 2) the NSRM generates an action $a_t^d$  by policy $\pi^d$ to choose the depth of subgraph, then the policy $\pi^w$ generates an action $a_t^w$ to choose the width of subgraph.}
\label{fig_agent_flow}
\end{figure*}

\subsection{Critical Brain Subgraph Mining}
\label{sec:mining}
To adaptively mine subgraphs of arbitrary size and shape without supervision, 
we develop two modules: the Loop Subgraph Reinforced Mining module (LSRM) and the Neighbor Subgraph Reinforced Mining module (NSRM). Both are designed to model the subgraph mining approach as a Finite Horizon Markov Decision Process (MDP), enabling autonomous discovery of critical subgraphs guided by downstream task feedback.
As depicted in Fig.\ref{fig_agent_flow}, 
our method begins by selecting a set of seed nodes as the central nodes of the subgraph.
LSRM with its loop agent, mines loop subgraphs that contain $n_i$, facilitating the analysis of discriminative connectivity information. 
Concurrently, NSRM with its depth and width agents, mines the neighbor subgraphs, focusing on local neighborhoods to gather information.
These two approaches allow for a comprehensive analysis of both loop structures and neighboring interactions, enhancing our understanding of brain connectivity patterns.

\begin{figure}[!ht]
\centering
\includegraphics[width=5.5in]{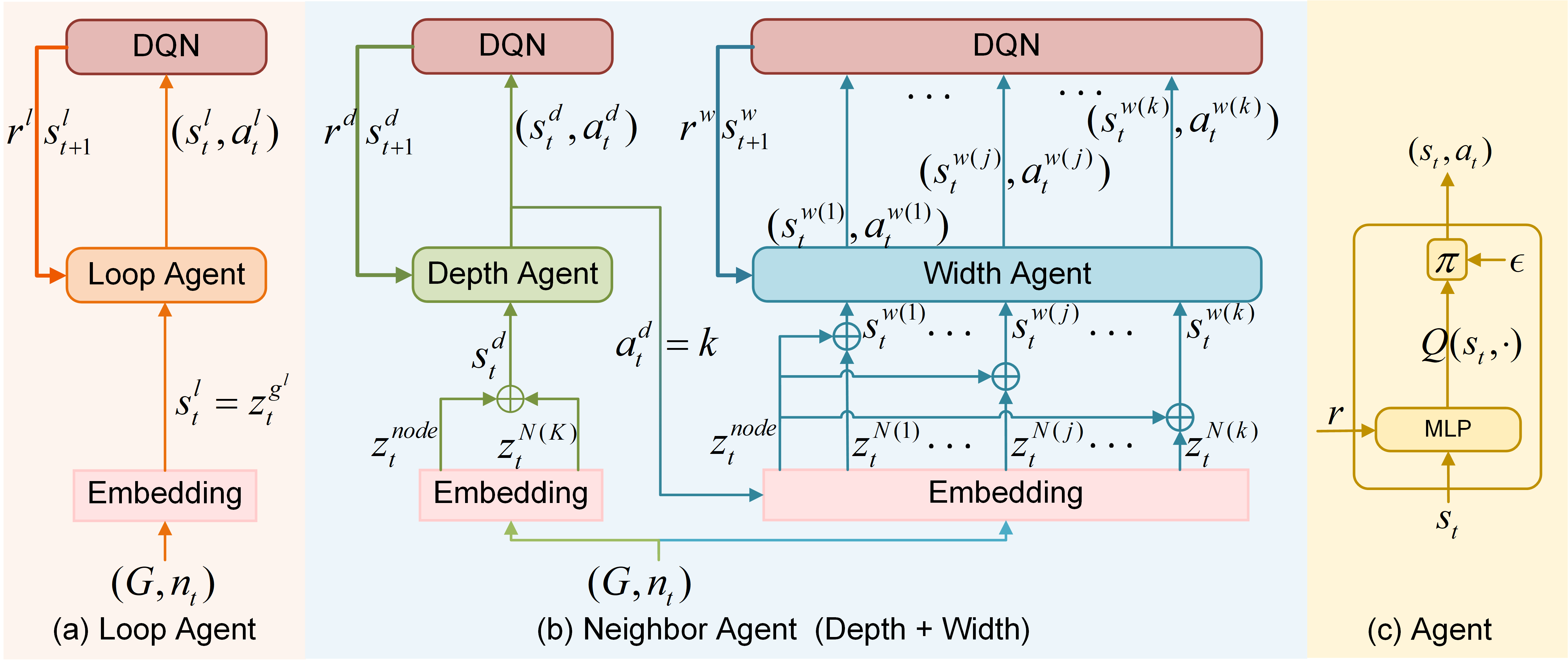}%
\caption{The agents of subgraph mining module.}
\label{fig_agent}
\end{figure}

\subsubsection{\textbf{Loop Subgraph Reinforced Mining Module}}
We first sample a set of seed nodes $V_{seed}=\{n_i\}_{i=1}^M$ as central nodes of the subgraph, and then pick the graph around these seed nodes. 
For each central node $n_i$, an encoder initially transforms node features into node codes. 
The loop subgraphs containing $n_i$ are then mined with a loop agent.

The loop agent operates by generating a series of actions $\{a_t^{l(j)}| j\in {[0,\delta]}\quad and\quad j\in \mathbb{Z}\}$ to select loop subgraphs $g^l$ according to the policy $\pi^l(s_t^l)$, where $\delta$ denotes the toal number of subgraphs. 
Specifically, the loop agent is 
defined by four components
$(\mathcal{S}^l,\mathcal{A}^l,\mathcal{R}^l,\mathcal{T}^l)$: 
1) \textbf{S}tate $\mathcal{S}^l$ consists of the embedding of all nodes of the currently selected subgraph, denoted as $s_t^l=\{z^{g^l}_t\}$, where $g^l$ is the currently selected subgraph;
2) \textbf{A}ction $\mathcal{A}^l$ consists of a series of subgraphs selecting the action $\{a_t^{l(j)}| j\in {[0,\delta]}\quad and\quad j\in \mathbb{Z}\}$ for selecting the current subgraph $g_t^l$ and $\delta$ representing the number of all subgraphs;
3) \textbf{R}eward $\mathcal{R}^l$ is defined as a discrete reward function $r(s_t^l,a_t^l)$, 
since it is unavailable to determine the state and its accumulated reward in GNN after each action is taken. As illustrated in (\ref{r^l}), reward determineswhether the reward of $(s_t^l,a_t^l)$ is positive based on whether the predicted label $\hat{y}_t$ is equal to the true label $y$;
4) \textbf{T}ransition $\mathcal{T}^l$ samples the next subgraph after the current subgraph is determined, until $\delta$ subgraphs are sampled.
\begin{equation}
r(s_t^*,a_t^*)=
\begin{cases}
+0.5, & \text { if } \hat{y}_t=y \\ -0.5, & \text { if } \hat{y}_t \neq y
\end{cases}
\label{r^l}
\end{equation}
After completing the setup of loop agent, to achieve the overall steps of maximizing the expected total reward by starting from the current state, we apply the model-free algorithm DQN to learn the optimal policy for the agents, which is completely alternative. Specifically, DQN learns the state-action values $Q^l\left(s^l, a^l\right)$ by using a deep neural network, we therefore train $Q^l$ functions for loop agent based on (\ref{Q^l}):
\begin{equation}
\begin{aligned}
Q^l\left(s^l, a^l\right) =\mathbb{E}_{s^{l^{\prime}}}\left[\mathcal{R}\left(s^{l^{\prime}}\right)+\gamma^l \max _{a^{l^{\prime}}}\left(Q^l\left(s^{l^{\prime}}, a^{l^{\prime}}\right)\right]\right.
\end{aligned}
\label{Q^l}
\end{equation}
where $\gamma^l\in[0,1]$ is the discount factors of future reward. Then the policies $\pi^l$ are obtained by a $\epsilon$-greedy policy with an explore probability $\epsilon$ as:
\begin{equation}
\begin{aligned}
& \pi^l\left(a_t^l \mid s_t^l ; Q^{l *}\right)=\left\{\begin{array}{cc}
\text { random action, } & p=\epsilon \\
\underset{a_t^l}{\arg \max } Q^{l *}\left(s_t^l, a_t^l\right), & p=1-\epsilon
\end{array}\right.
\end{aligned}
\end{equation}

\begin{equation}
\pi^l(G)\Rightarrow g_j^l
\label{pi^l}
\end{equation}
where $\pi^l$ is used to explore new states by choosing random actions with probability $\epsilon$, $p$ is the probability of current action. Briefly, we first mine all the loop subgraphs from the original graph $G$ and input them to $\pi^l$ one by one as (\ref{pi^l}) to further analyze discriminative brain connectivity information.
Notably, the brain loops here are not clinically disorder in function but are closed connections connected by disorder task-related brain regions. Besides, some graphs may not contain loop subgraphs.

\subsubsection{\textbf{Neighbor Subgraph Reinforced Mining Module}}
To mine subgraphs of arbitrary size and shape, we split the NSRM module into depth and width agents to mine the appropriate depth and width for subgraph. Specifically, for node $n_t$ in state $\mathcal{s}_{t}$, the depth agent generates an action $a_t^d$ according to the policy $\pi_t^d(*)$ to define the depth $h_t$ for the subgraph of node $n_t$. When $h_t=a_t^d$, the width agent will generate a series of actions $\{a_t^{w(i)}| i\in {[1,h]}\quad and\quad i\in \mathbb{Z}\}$ for sampling the member nodes of subgraph within k-hop neighbors of $n_t$ hop by hop. Our two agents are described in detail below.

The depth agent is denoted a tuple $(\mathcal{S}^d,\mathcal{A}^d,\mathcal{R}^d,\mathcal{T}^d)$:
1) \textbf{S}tate $\mathcal{S}^d$ is denoted as $s_t^d=(z^{node}_t,z^{N^{(K)}}_t)$, comprising the embedding $z^{node}_t$ of current central node $n_t$ and the context embedding $z^{N^{(K)}}_t$ of its k-hop neighbors. The $z^{N^{(K)}}_t$ is the average vector of all nodes in the k-hop neighborhood of current node $n_t$, and $K$ is the hyperparameter that limits the maximum subgraph depth;
2) \textbf{A}ction $\mathcal{A}^d$ consists of a series of depth selection actions $\{a_t^d\}$ for selecting the maximum depth of the subgraph from node $n_t$, where $a_t^{d(i)}\in[1,K]$ is a positive integer;
3) \textbf{R}eward $\mathcal{R}^d$ is defined as a discrete reward function $r(s_t^d,a_t^d)$ illustrated in (\ref{r^l}), the reward of $(s_t^d,a_t^d)$ is positive when the predicted label $\hat{y}_t$ is equal to the true label $y$;
4) \textbf{T}ransition $\mathcal{T}^d$ samples the next center node from the seed nodes as the next state $s_{t+1}^d$ once the depth of the current subgraph is determined, until $N$ subgraphs are sampled.

The width agent is denoted as $(\mathcal{S}^w,\mathcal{A}^w,\mathcal{R}^w,\mathcal{T}^w)$ in a similar way:
1) \textbf{S}tate $\mathcal{S}^w$ is denoted as a sires of $\{s_t^{w(k)}\}$, which consists of the spatial embedding $z^{g_{k-1}}_t$ of the subgraph of the current center node $n_t$ and the context embedding $z^{N^{(K)}}_t$ in the $n_t$'s $k$-th hop. The $z^{g_{k-1}}_t$ is the average vector of all nodes in the current $(k-1)$-hop subgraph, and $z^{g_1}$ is initialized to the embedding representation of the central node $n_t$, $z^{N^{(K)}}_t$ is the average vector of all nodes in the $k$-th hop neighbors of node $n_t$;
2) \textbf{A}ction $\mathcal{A}^w$ consists of a series of width selection actions $a_t^w=\{a_t^{w(i)}| i\in {[1,k]}\quad and\quad i\in \mathbb{Z}\}$ when $k=a_t^d$, where $a_t^{w(i)}$ represents the action in $i$ -th hop for width sampling;
3) \textbf{R}eward $\mathcal{R}^w$ is same as $\mathcal{R}^d$, illustrated in (\ref{r^l});
4) \textbf{T}ransition $\mathcal{T}^d$ represents the current subgraph and the next hop node as the next state $s_{t+1}^w$ after the current subgraph depth is determined, until all k-hop nodes of the subgraph are completed sampling.

After completing the setup of both agents, we apply the DQN to learn the state-action values $Q^d$ and $Q^w$ (Here denoted uniformly as $Q$) by using a deep neural network for depth and width agents:
\begin{equation}
Q\left(s, a\right) =\mathbb{E}_{s^{\prime}}\left[\mathcal{R}\left(s^{\prime}\right)+\gamma \max _{a^{\prime}}\left(Q\left(s^{\prime}, a^{\prime}\right)\right]\right.
\end{equation}
where $\gamma^d\in[0,1]$ and $\gamma^w\in[0,1]$ is the discount factors of future reward. Then the policies $\pi^d$ and $\pi^w$ are obtained by a $\epsilon$-greedy policy with an explore probability $\epsilon$ as:
\begin{equation}
\pi\left(a_t \mid s_t ; Q^{*}\right)=\left\{\begin{array}{cc}
\text{random action},  & p=\epsilon \\
\underset{a_t}{\arg \max } Q^{*}\left(s_t, a_t\right), & p=1-\epsilon
\end{array}\right.
\end{equation}
where $\pi^d$ and $\pi^w$ are used to explore new states by choosing random actions with probability $\epsilon$. Briefly, as (\ref{pi^d}), we first input the original graph $G$ and nodes $n_i$ to $\pi^d$ to get the subgraph depth optimum $d_i$, and then input $G$, $n_i$ and $d_i$ to $\pi^w$ to get the final selected nodes of the subgraph, thus obtaining the neighborhood subgraph representing the neighborhood context information.
\begin{equation}
    \pi^d(G,n_i)\Rightarrow d_i,\quad    \pi^w(G,n_i,d_i)\Rightarrow g_i
    \label{pi^d}
\end{equation}

\subsection{Optimization Process}
To achieve optimized results, we unify the training of brain network construction and subgraph mining. Specifically, optimization process contains: 1) pretraining the backbone to initialize feature extraction; 2) clustering feature channels with K-means; 3) pretraining CCL to align with the k-means operation; 4) freezing the subgraph mining enhancement module of graph convolution, training the combined backbone, CCL and Brain-SubGNN to encode and optimize the brain network graph; 5) jointly training all modules of Brain-SubGNN for subgraph mining and enhancement. The overall loss function is defined as:
\begin{equation}
\mathcal{L}=\mathcal{L}_{ccl}+\mathcal{L}_{g}+\beta\mathcal{L}_{MI}
\label{loss}
\end{equation}
where $\mathcal{L}_{ccl}$ is the channel clustering loss described in (\ref{L_ccl}), $\mathcal{L}_g$ is the cross-entropy loss for global graph classification, $\mathcal{L}_{MI}$ is the MI loss between the subgraph and the global graph, and $\beta$ is the hyper-parameter controlling the impact of $\mathcal{L}_{MI}$.

\begin{table*}[!t]
\caption{Demographic information of the subjects. The gender is presented as male/female, Age and education years are presented as mean ± standard deviation}
  \centering
    \begin{tabular}{cccccc}
    \hline\hline
    \textbf{Dataset} & \textbf{Type}  & \textbf{Num} & \textbf{Gender(M/F)} & \textbf{Age(year)} & \textbf{Education} \\
    \hline
    \multirow{4}{*}{ADNI-1} & sMCI  & 197   & 126/71 & 75.03±7.36 & 15.59±3.10 \\
          & pMCI  & 108   & 68/40 & 77.59±7.57 & 15.73±2.61 \\
          & sNC   & 249   & 121/127 & 79.72±7.07 & 16.55±2.62 \\
          & pNC   & 80   & 47/33 & 82.96±5.77 & 16.02±2.58 \\
    \hline
    \multirow{2}{*}{ADNI-2} & sMCI  & 251   & 138/113 & 71.46±7.31 & 16.32±2.70 \\
          & pMCI  & 99    & 55/44 & 76.72±6.39 & 16.05±2.66 \\
    \hline
    \multirow{2}{*}{NACC}  & sNC   & 281   & 74/207 & 66.94±8.60 & 16.00±2.39 \\
          & pNC   & 187   & 64/123 & 75.44±10.34 & 16.34±9.15 \\
    \hline
    \multirow{2}{*}{AIBL}  & sNC   & 59   & 36/23 & 75.63±6.38 & - \\
          & pNC   & 30   & 20/10 & 73.63±6.41 & - \\
    \hline\hline
    \end{tabular}%
  \label{tab_dataset}%
\end{table*}%

\section{Experiments and Results}

\subsection{Materials and Preprocessing}
The data utilized in this study are from three publicly available database: Alzheimer’s Disease Neuroimaging Initiative 1 and 2 (ADNI-1, ADNI-2) \cite{petersen2010alzheimer}, National Alzheimer’s Coordinating Center (NACC) \cite{besser2018version}, including 1.5T, 2T and 3T T1-weighted MRI data. We collect 1123 subjects in all datasets, including 448 sMCI, 207 pMCI, 281 sNC, and 187 pNC. Note that for subjects appearing in multiple datasets of ADNI (e.g., ADNI-1 and ADNI-2), we only keep the former. Demographic details of the subjects are shown in Tab.~\ref{tab_dataset}.

We preprocess the T1-MRI data as follows: 1) motion correction and conform; 2) non-uniform intensity normalization using the N3 algorithm; 3) computation of Talairach transform; 4) intensity normalization; 5) skull stripping and affine registration using FreeSurfer (https://fsl.fmrib.ox.ac.uk/); 6) spatial normalization to the Montreal Neurological Institute (MNI) space with a resolution of 3mm × 3mm × 3mm using the Statistical Parametric Mapping (SPM); and 7) smoothing with an 8 mm kernel in MATLAB 2020a. The resulting images are resized to 91 × 109 × 91 voxels.

\subsection{Implementation Details}
\label{implementation}
We choose convimixer \cite{trockman2022patches} as the backbone with convolutional kernel=5, depths=5 and channels=2048. K-means is set with cluster centers $N_c$=16, and the CCL is a two-layer FCs implementing prototype contrast learning with smooth parameter $\alpha$=10 following \cite{chen2020improved}. The Brain-SubGNN consists of a two-layer DenseGCN \cite{li2021deepgcns}, a two-layer Graph Isomorphism Network (GIN) \cite{xu2018powerful}, two subgraph mining module to implement original graph encoding, subgraph encoding, critical subgraph mining and subgraph enhancement, respectively. The maximum depth of neighbor subgraph=3, the number of mined subgraphs is chosen according to the size of the graph. The $\beta$ of MI loss is set as 0.8. The $Q$ function of DQN is set to a 5-level MLP with (128, 256, 512, 256, 128) hidden units, and the $\epsilon$ of the $\epsilon$-greedy policy decreases from 1.0 to 0.2 at a rate of 0.95 per 100 steps according to a linear scheduler.

The model is trained for 300 epochs with initial learning rates of $1e^{-4}$, $1e^{-6}$ and $1e^{-2}$ for the backbone, CCL and Brain-SubGNN, respectively. The learning rates are decreased by a factor of 10 every 100 epochs. Evaluation metrics include accuracy (ACC), sensitivity (SEN), specificity (SPE), and area under the curve (AUC). We use Python and PyTorch package, and run the network on a single NVIDIA GeForce 3090 GPU.

Our model is evaluated on two tasks, NC conversion prediction (sNC vs. pNC) and MCI conversion prediction (sMCI vs. pMCI). 
As the NC data is derived from a single dataset and the MCI data comes from two distinct datasets, we therefore perform a 5-fold cross-validation on sNC vs. pNC and  sMCI vs. pMCI tasks in all experiments, and an independent validation on sMCI vs. pMCI task for SOTA comparison.

\begin{table*}[!bhtp]
  \centering
  \caption{Comparsion of our method with current SOTA methods for MCI conversion
prediction on ADNI-2, obtained by the models trained on ADNI-1}
    \begin{tabular}{ccccrrrr}
    \hline\hline
    \multicolumn{1}{c}{Method} & Cite & \makecell[c]{ADNI-1 \\ (sMCI/pMCI)} & \makecell[c]{ADNI-2 \\ (sMCI/pMCI)}& ACC & SEN & SPE & AUC \\
    \hline
    LDMIL & \cite{liu2018landmark}  & 226/167 & 239/38 & 0.769  & 0.421  & 0.824  & 0.776  \\
    H-FCN & \cite{lian2018hierarchical} & 226/167 & 239/38 & 0.809  & 0.526  & 0.854  & 0.781  \\
    HybNet & \cite{lian2020attention} & 226/167 & 239/38 & 0.827  & 0.579  & 0.866  & 0.793  \\
    AD$^2$A & \cite{guan2021multi} & 147/165 & 253/88 & 0.780  & 0.534  & 0.866  & 0.788  \\
    DSNet & \cite{pan2021disease}  & 147/165 & 256/89 & 0.762  & \textbf{0.770} & 0.742  & 0.818  \\
    MSA3D & \cite{chen2022alzheimer} & 229/167 & 241/75 & 0.801  & 0.520  & 0.856  & 0.789  \\
    M$^2$FAN & \cite{han2023multi} & 165/175 & 323/109 & 0.815 & 0.670 & 0.864 & 0.817 \\
    DH-ProGCN & \cite{leng2023dynamic} & 197/108 & 251/99 & 0.849  & 0.647  & \textbf{0.928} & 0.845  \\
    \textbf{Ours} & -  & 197/108 & 251/99 & \textbf{0.869} & 0.727  & 0.924  & \textbf{0.882} \\
    \hline\hline
    \end{tabular}%
  \label{tab_sota}%
\end{table*}%

\subsection{Comparison with the State-of-the-Art Methods}
After in-depth survey, we find there are no Deep Learning study on sNC vs. pNC classification using T1-MRI images. Therefore, we here focus on presenting SOTA comparison for sMCI vs. pMCI classification.
Eight SOTA methods are used for comparison: 
1) LDMIL \cite{liu2018landmark} captured both local information conveyed by patches and global information; 
2) H-FCN \cite{lian2018hierarchical} implemented three levels of networks to obtain multi-scale feature representations which are fused for the construction of hierarchical classifiers; 
3) HybNet \cite{lian2020attention} assigned the subject-level label to patches for local feature learning by iterative network pruning; 
4) AD$^2$A \cite{guan2021multi} located discriminative disease-related regions by an attention modules;
5) DSNet \cite{pan2021disease} provided disease-image specificity to an image synthesis network; 
6) MSA3D \cite{chen2022alzheimer} implemented a slice-level attention and a 3D CNN to capture subject-level structural changes;
7) M$^2$FAN \cite{han2023multi} designed modules of weakly supervised meta-information learning to learn and disentangle meta-information from class-related representations;
8) DH-ProGCN \cite{leng2023dynamic} constructed hierarchical structure brain networks for classification.

The comparisons between Brain-SubGNN and existing SOTA methods are presented in Tab.~\ref{tab_sota}. Notably, Brain-SubGNN demonstrates superior performance in MCI conversion prediction, achieving ACC=0.869 and AUC=0.882 in tests conducted on the ADNI-2, with models trained on ADNI-1.

\begin{table}[thb]
  \centering
  \caption{Effects of different types of nodes and edges}
    \begin{tabular}{c|cc|cccc}
    \hline\hline
          & \multicolumn{2}{c|}{\textbf{dynamic}} & \multicolumn{4}{c}{\textbf{5-fold validate on ADNI-1\&2}} \\
          & node & edge & ACC   & SEN   & SPE   & AUC \\
    \hline
    sMCI  & \multicolumn{2}{c|}{ROI-based} & 0.774 & 0.606 & 0.841 & 0.793 \\
     vs.  &      &      & 0.798 & 0.695 & 0.804 & 0.803 \\
    pMCI  & \checkmark     &      & 0.800 & 0.668 & 0.854 & \textbf{0.831} \\
          & \checkmark     & \checkmark     & \textbf{0.810} & \textbf{0.696}  & \textbf{0.857} & \textbf{0.831} \\
    \hline
    sNC   & \multicolumn{2}{c|}{ROI-based} & 0.847 & 0.828 & 0.854 & 0.916 \\
    vs.   &      &      & 0.851 & 0.816 & 0.875 & 0.921 \\
    pNC   & \checkmark     &      & 0.870 & 0.840 & 0.890 & \textbf{0.927} \\
          & \checkmark     & \checkmark     & \textbf{0.880} & \textbf{0.861} & \textbf{0.893} & 0.922 \\
    \hline\hline
    \end{tabular}%
  \label{tab_ablation_dynamic}%
\end{table}%

\subsection{Effectiveness of Dynamic Brain Network Construction}
\subsubsection{Dynamic critical region exploring}
To validate the effectiveness of dynamic critical region exploration, we compared 
1) the ROI-based approach, choose 90 regions of AAL as nodes;
2) the backbone network without channel clustering; 
and 3) the backbone network with channel clustering (i.e., dynamic node). As shown in Tab.~\ref{tab_ablation_dynamic}, the dynamic clustering outperforms the ROI and backbone based approaches in terms of MCI conversion and can produce better feature distributions for downstream brain image analysis tasks.

\begin{table*}[bthp]
  \centering
  \caption{Effects of different subgraph mining and enhancement strategies}
    \begin{tabular}{c|ccc|cccc}
    \hline\hline
    & \multicolumn{3}{c|}{\textbf{strategies}} & \multicolumn{4}{c}{\textbf{5-fold validate on ADNI-1\&2}} \\
    & Neighbor & Loop & MI    & ACC   & SEN   & SPE   & AUC \\
    \hline
    &      &      &      & 0.810  & 0.696  & 0.857  & 0.831  \\
    & \checkmark     &      &      & 0.835  & 0.674  & 0.902  & 0.842  \\
    sMCI & \checkmark     &      & \checkmark     & 0.845  & 0.662  & 0.920  & 0.849  \\
    vs. &      & \checkmark     &      & 0.836  & 0.640 & 0.916  & 0.846  \\
    pMCI &     & \checkmark     & \checkmark     & 0.850  & 0.636  & \textbf{0.938}  & 0.855  \\
    & \checkmark     & \checkmark     &      & 0.839  & 0.644  & 0.918  & 0.851  \\
    & \checkmark     & \checkmark     & \checkmark     & \textbf{0.858} & \textbf{0.717}  & 0.936 &  \textbf{0.860} \\
    \hline
    &      &      &      & 0.880  & 0.861  & 0.893  & 0.922  \\
    & \checkmark     &      &      & 0.889  & 0.847  & 0.918  & 0.934  \\
    sNC & \checkmark     &      & \checkmark     & 0.904  & \textbf{0.889}  & 0.914  & 0.936  \\
    vs. &      & \checkmark     &      & 0.882  & 0.824  & 0.922  & 0.929  \\
    pNC &      & \checkmark     & \checkmark     & 0.894  & 0.868  & 0.911  & 0.895  \\
    & \checkmark     & \checkmark     &      & 0.896  & 0.878  & 0.908  & 0.921  \\
    & \checkmark     & \checkmark     & \checkmark     & \textbf{0.917} &  0.874 & \textbf{0.947} & \textbf{0.949} \\
    \hline\hline
    \end{tabular}%
  \label{tab_ablation_subgraph}%
\end{table*}%

\subsubsection{Dynamic edge connection}
To verify whether our constructed dynamic brain network outperforms the fixed structure, we directly connect all critical region nodes after channel clustering and input them into a two-layer GNN for classification to obtain a fixed brain network map. The results are shown in Tab.~\ref{tab_ablation_dynamic}, where the dynamic brain network structure performs better, indicating that dynamic measurement of inter-regional correlations is necessary to construct the brain network.

\subsection{Effectiveness of Critical Subgraph Mining}
\subsubsection{subgraph mining}
The results of the ablation experiments for subgraph mining on two tasks are summarized in rows 2, 4 and 6 of Tab.~\ref{tab_ablation_subgraph}. 
The experiments reveal a sequential increase in average accuracy from the baseline to the incorporation of loop and neighbor subgraph mining modules, both individually and in combination.
This consistent improvement supports 
the effectiveness of our critical subgraph mining strategy, emphasizing the Subgraph Reinforced Mining Module's role in improving graph classification by extracting more meaningful and predictive subgraphs.
\subsubsection{subgraph representation enhancement}
Rows 3, 5 and 7 of Tab.~\ref{tab_ablation_subgraph} summarizes the results of the ablation studies focusing on the subgraph representation enhancement module for two tasks. These results with MI enhancement in subgraph mining modules are better than those without it.

\begin{figure}[!t]
\centering
\subfloat[sMCI vs. pMCI]{\includegraphics[width=2.5in]{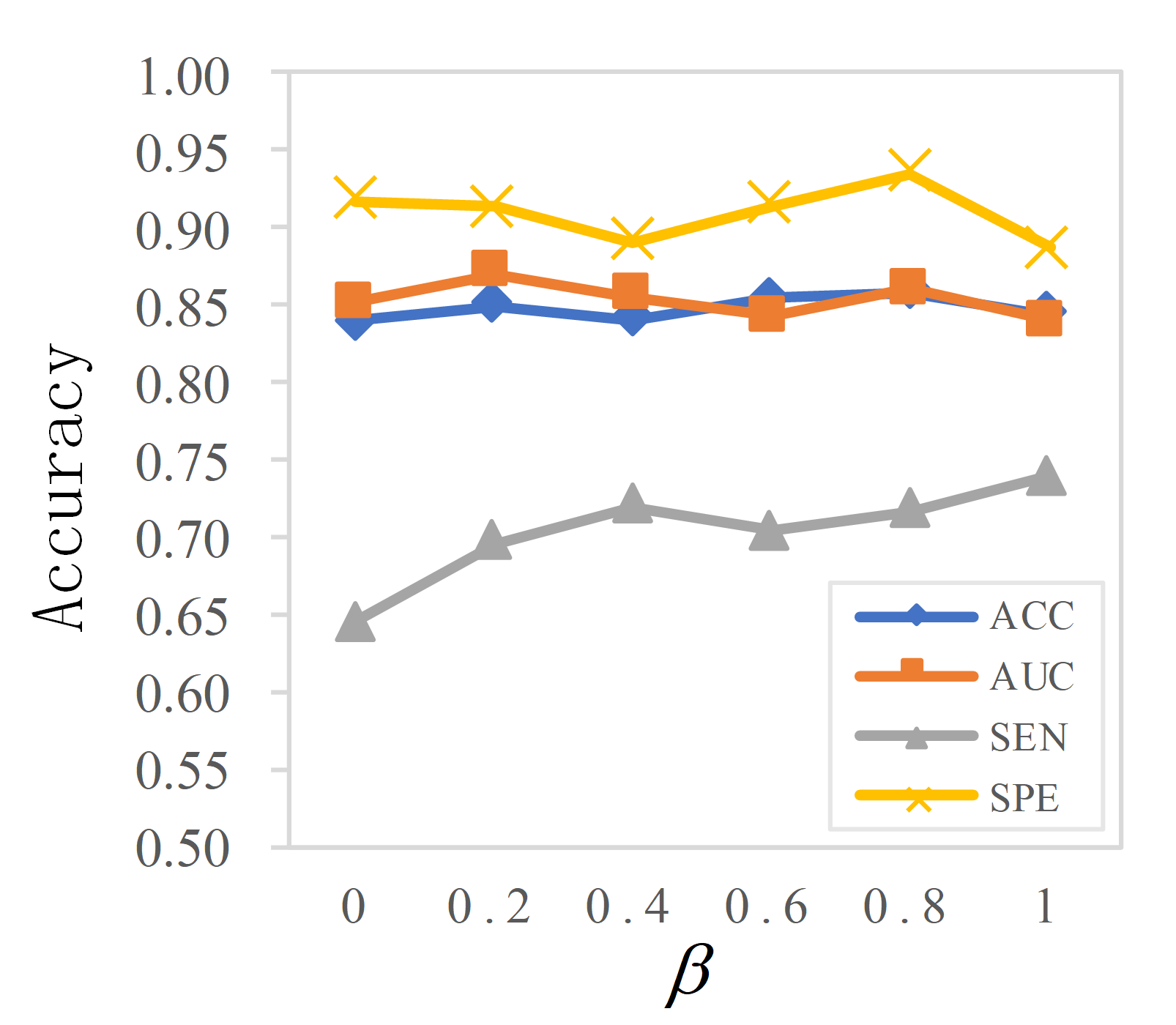}%
\label{fig_Ablation_MI_mci}}
\hfil
\subfloat[sNC vs. pNC]{\includegraphics[width=2.5in]{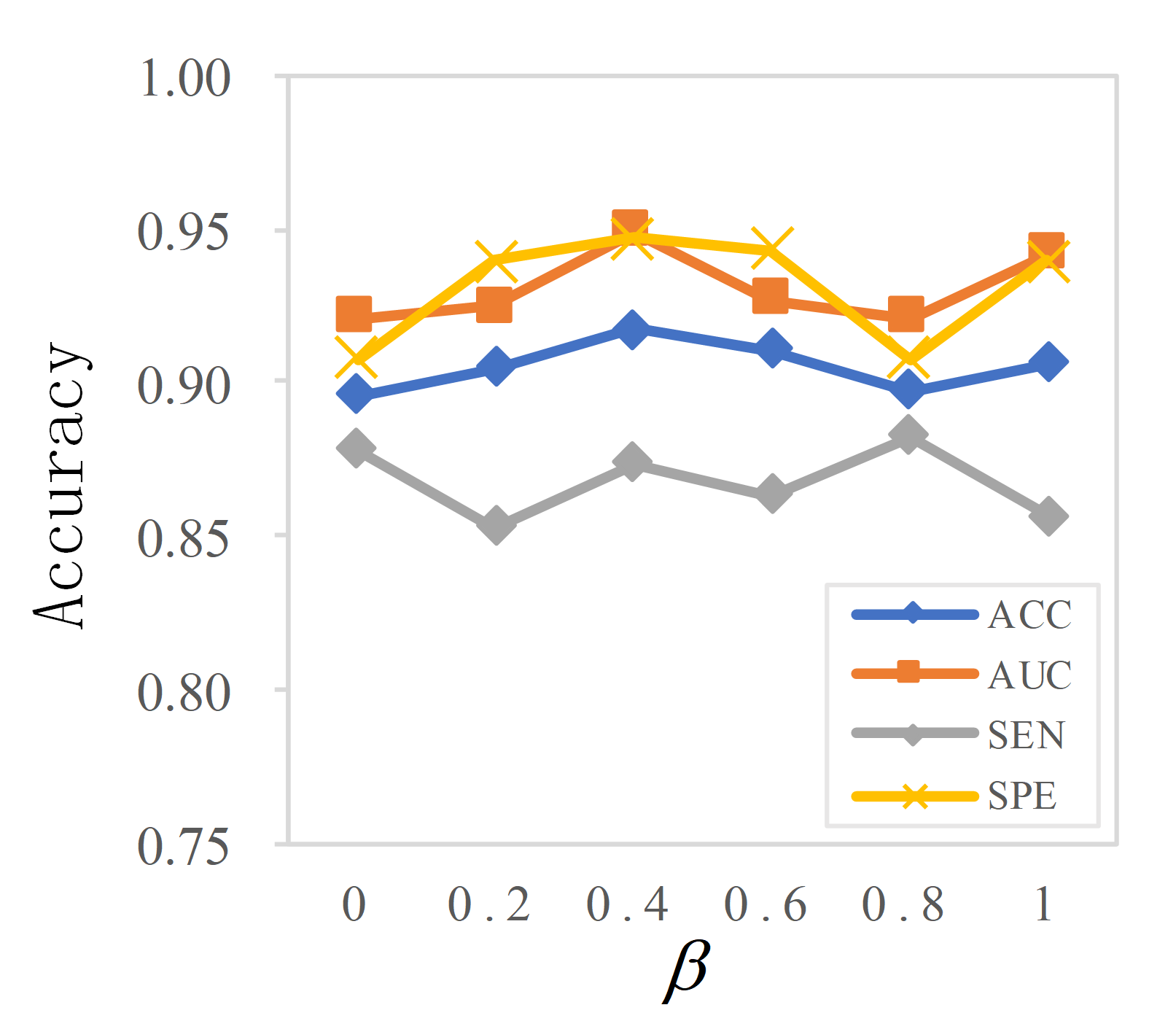}%
\label{fig_Ablation_MI_nc}}
\caption{Effects of the coefficient of MI module in controlling the contribution of the Global-Aware MI loss.}
\label{fig_Ablation_MI}
\end{figure}

\subsubsection{MI Coefficient}
The sensitivity of the MI loss coefficient $\beta$ is designed for controlling the contribution of the Global-Aware MI loss $L_{global}$, and analyzed across two datasets in (\ref{loss}).
As observed in Fig.\ref{fig_Ablation_MI}, Brain-SubGNN achieves optimal on both datasets when $\beta$ ranges 0.4 to 0.6. This indicates that 
embedding overall graph properties into subgraphs is particularly beneficial for brain network analysis.
The proposed MI enhancement mechanism is adept at capturing diverse critical information specific to different dataset types.

\begin{figure}[!t]
\centering
\subfloat[sMCI vs. pMCI]{\includegraphics[width=2.5in]{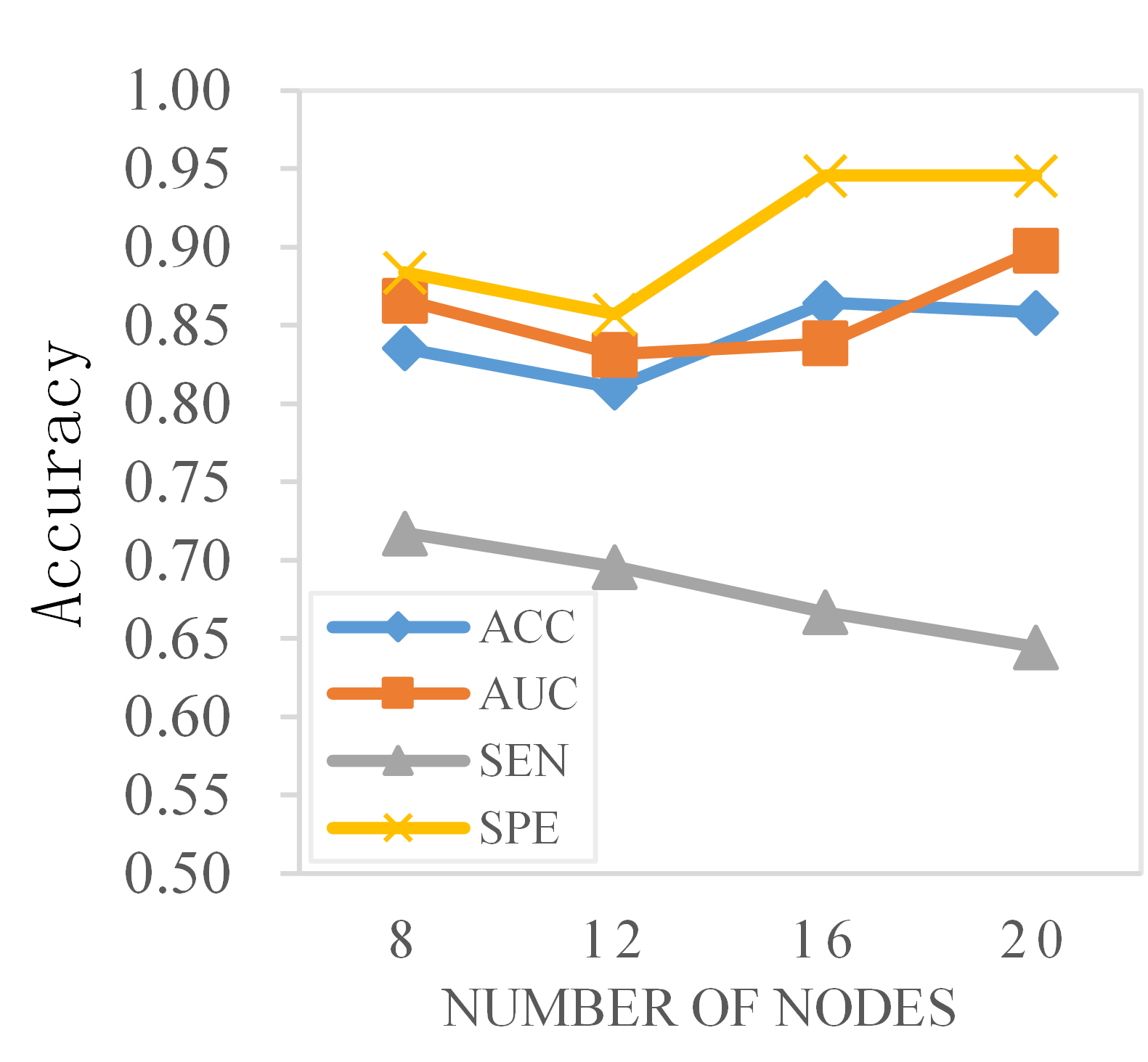}%
\label{fig_Ablation_N_mci}}
\hfil
\subfloat[sNC vs. pNC]{\includegraphics[width=2.5in]{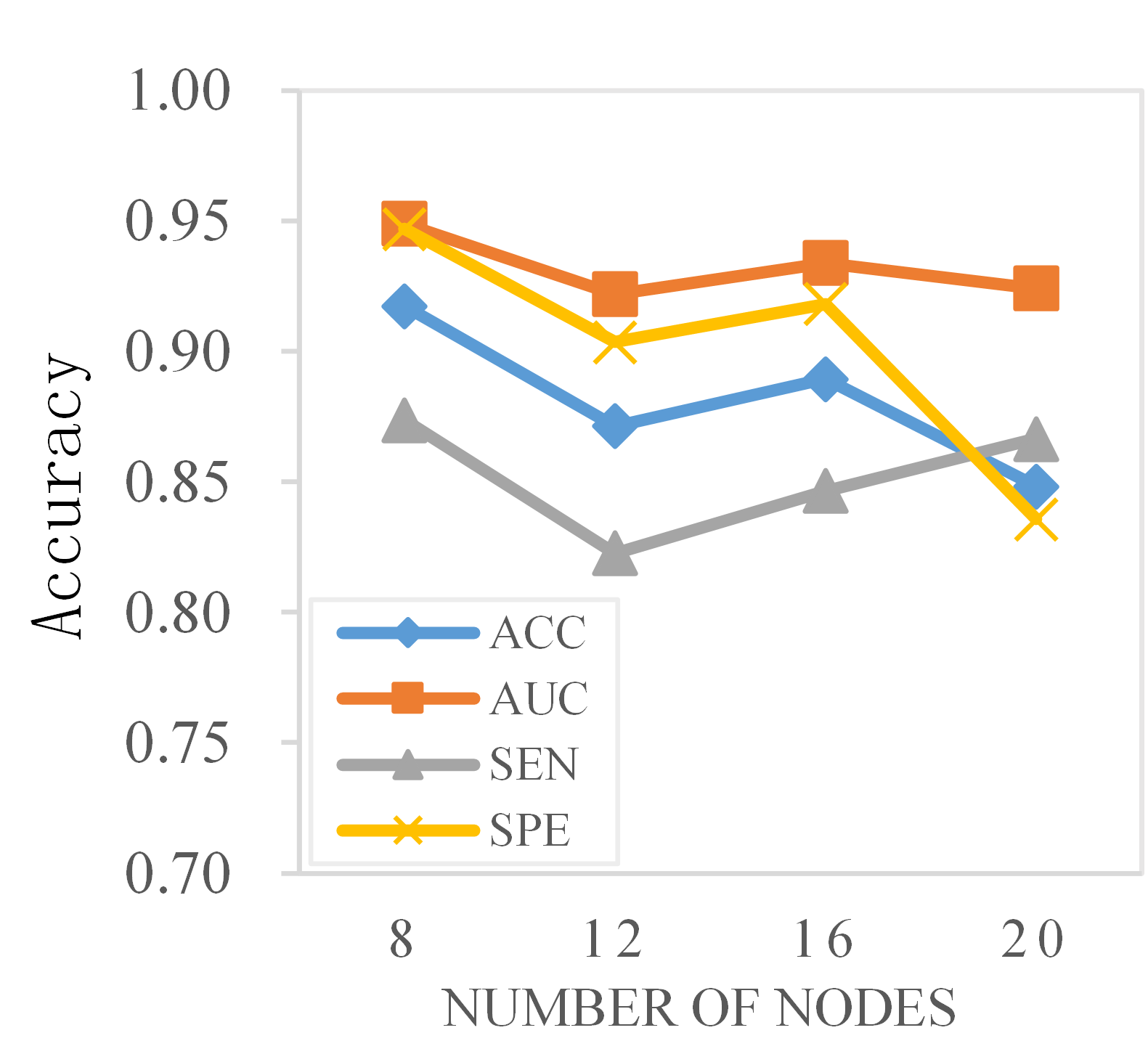}%
\label{fig_Ablation_N_nc}}
\caption{Effects of the number of nodes in constructing the brain network graph on the results with 5-fold cross-validated.}
\label{fig_Ablation_N}
\end{figure}

\begin{figure}[!t]
\centering
\subfloat[sMCI vs. pMCI]{\includegraphics[width=2.5in]{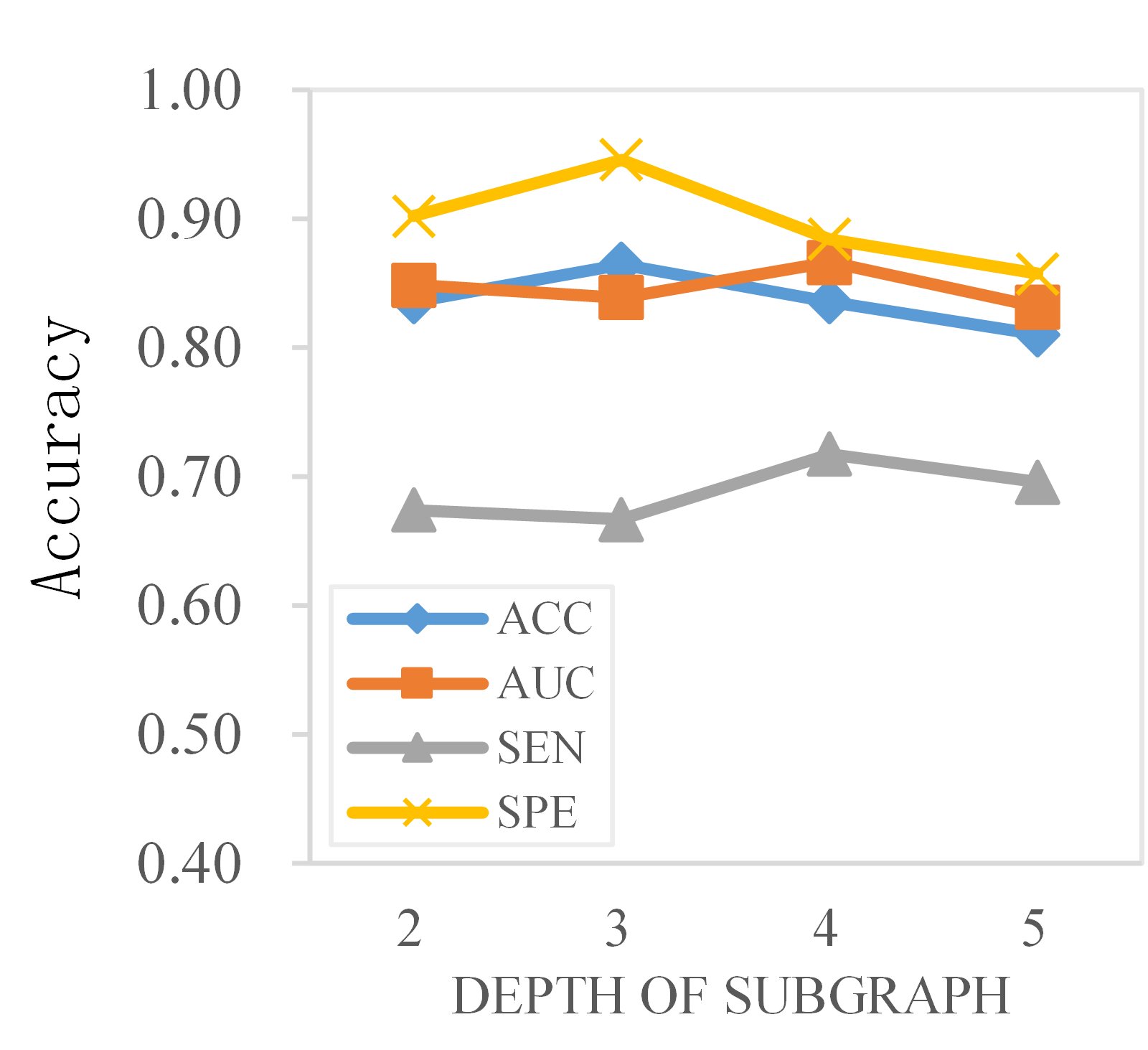}%
\label{fig_Ablation_Depth_mci}}
\hfil
\subfloat[sNC vs. pNC]{\includegraphics[width=2.5in]{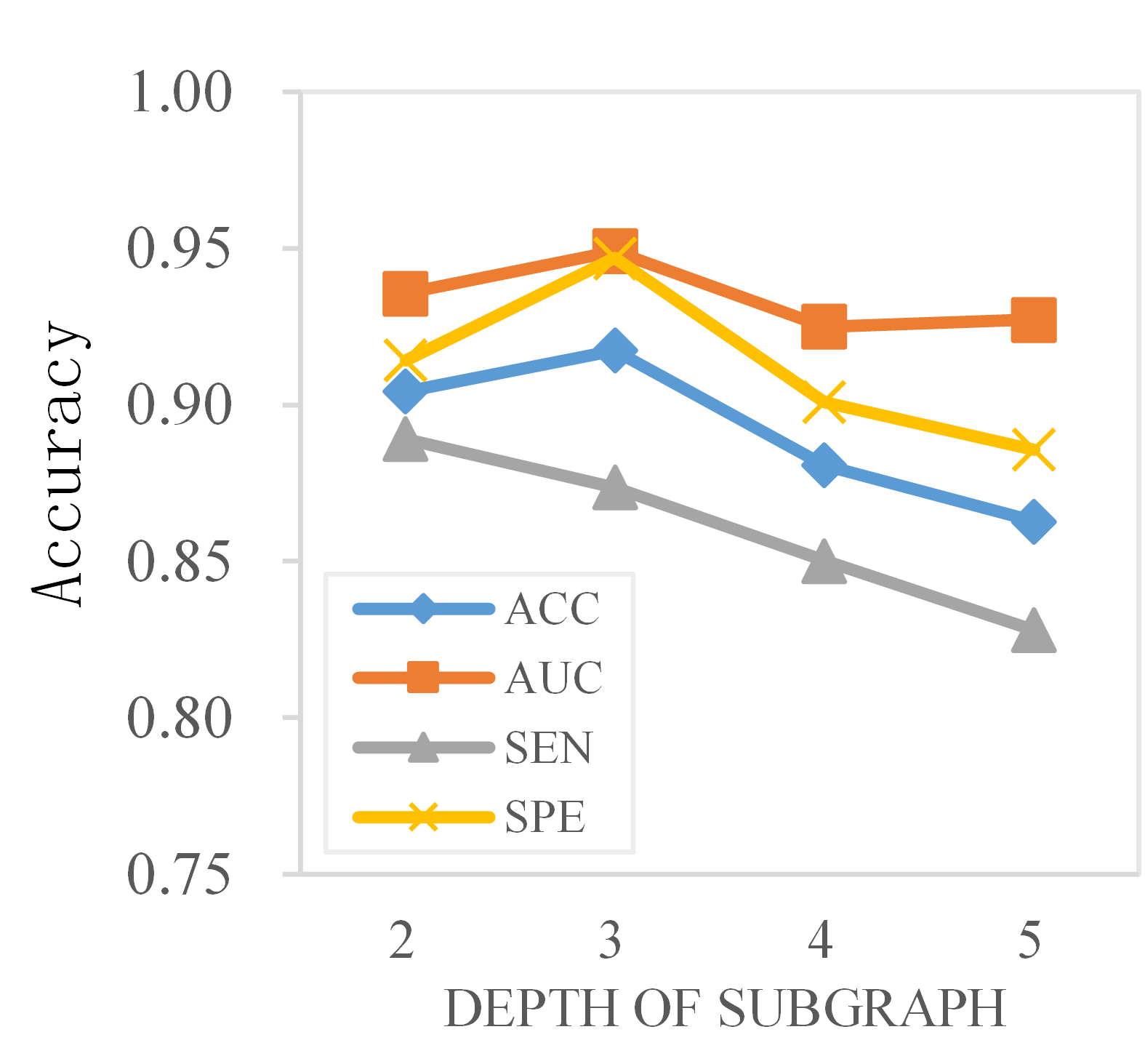}%
\label{fig_Ablation_Depth_nc}}
\caption{Effects of the depth of subgraph in mining the brain network representation with 5-fold cross-validated.}
\label{fig_Ablation_Depth}
\end{figure}

\begin{figure*}[!t]
\centering
\includegraphics[width=5.5in]{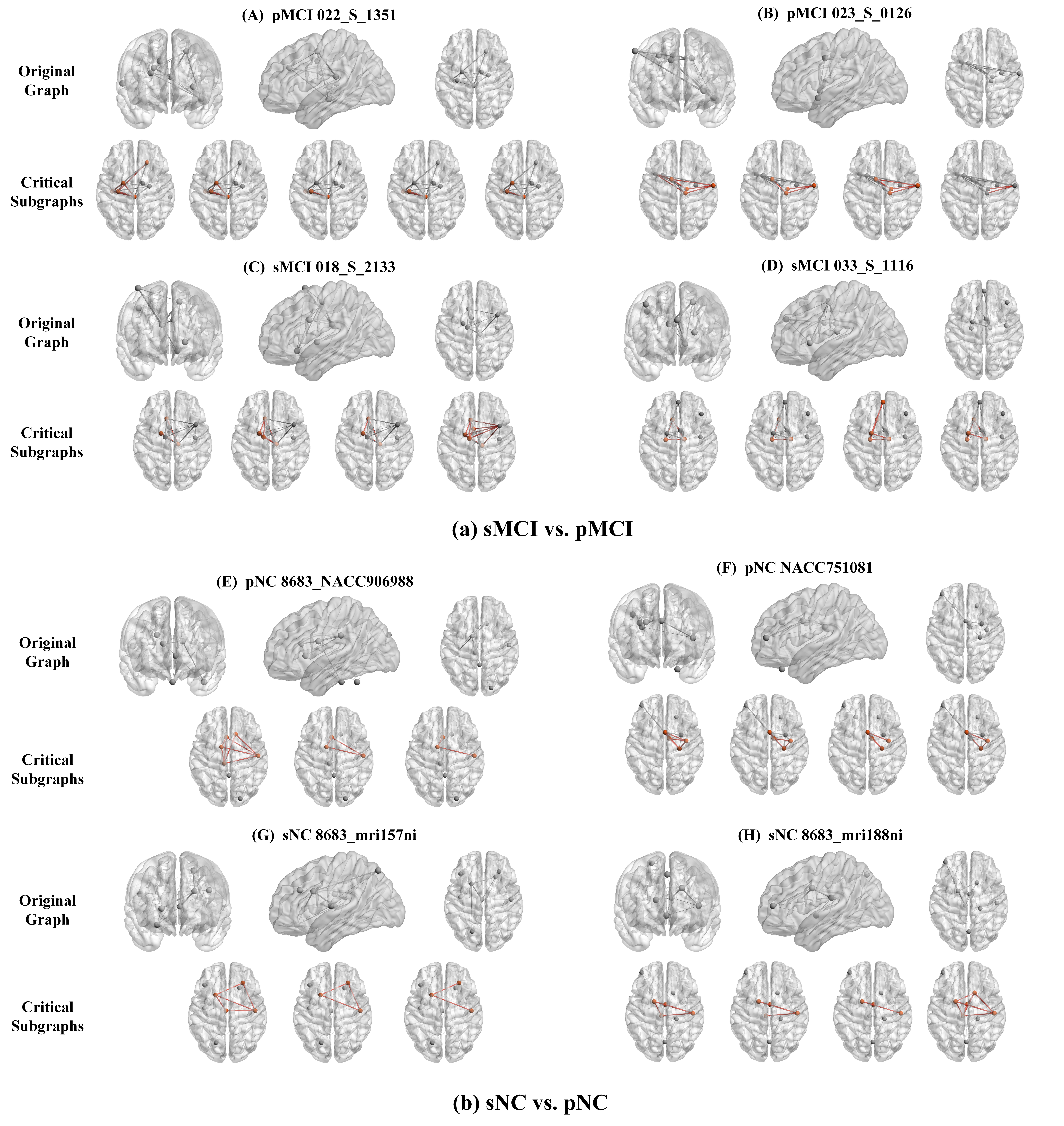}%
\caption{Visualization of the constructed original graphs and the mined critical subgraphs. (A)-(D) and (E)-(H) are the visualizations of MCI  and NC subjects, respectively. The gray nodes in the original graph are the discriminative regions by automatic localization, and the edges reflect their correlation. The red nodes and red edges are the components of critical subgraphs identified by Brain-SubGNN.}
\label{fig_subgraph}
\end{figure*}

\subsection{Effectiveness of Graph and Subgraph Representation}

\subsubsection{Size of Brain Graph}
To find the optimal graph representation of the brain network, we investigated how the size of the graph affects the performance of Brain-SubGNN. As described in Section \ref{sec:construct}, the number of nodes can be adjusted by controlling the number of channel clustering centers. The experimental results show that both too large and too small graph sizes degrade the performance of Brain-SubGNN. As shown in Fig.~\ref{fig_Ablation_N}, the best graph size of sMCI vs. pMCI and sNC vs. pNC (average number of vertices $N$=16) has the best performance compared with others. Too few nodes may over-generalize by encompassing excessive regions, thus limiting local specificity; whereas too many nodes might overly emphasize local variations, reducing the overall accuracy of the evaluation.
\subsubsection{Maximum Depth of Subgraph}
In Fig.~\ref{fig_Ablation_Depth}, we evaluate the proposed Brain-SubGNN with different maximum subgraph depths $K$ from 2 to 5 on sMCI vs. pMCI and sNC vs. pNC tasks. $K$ determines the agent state as well as the range of the action set. As shown in Fig.~\ref{fig_Ablation_Depth_mci} and Fig.~\ref{fig_Ablation_Depth_nc}, Brain-SubGNN achieves the best performance with $K$=3 in sMCI vs. pMCI and  sNC vs. pNC. This indicates that the critical subgraphs in MCI subject networks have a higher order than in NC subject networks.
This is logical, as pNC lesions are typically microlocalized, while MCI involves broader brain connectivity, affecting deterioration prediction. Hence, Brain-SubGNN excels when critical subgraphs encapsulate vital functional elements of the graph, balancing local and global analysis.

\subsection{Visualization}
We present a visualization of the brain networks and their corresponding critical subgraphs for both the sMCI vs. pMCI and sNC vs. pNC tasks in Fig.~\ref{fig_subgraph}. The original graphs are the brain networks constructed by Section.\ref{sec:construct}, while the critical subgraphs are mined by the subgraph mining module. 

Our framework identifies several critical regions and structures within the brain networks that are consistent with current neurological research.
For instance, in the case of sMCI vs. pMCI tasks, the critical subgraphs prominently feature regions such as the hippocampus and the entorhinal cortex \cite{du2001magnetic, pini2016brain}, which are known to be affected in the early stages of MCI and AD. 
Furthermore, our framework highlights a notable pattern of connectivity between the frontal lobe and the temporal lobe in the sNC vs. pNC tasks. 
The visualization reveals a decrease in connectivity in these regions in subjects progressing from sNC to pNC.
This observation echoes recent theories \cite{hampstead2023toward} suggesting the importance of frontal-temporal connections in the maintenance of cognitive functions, particularly in the context of normal aging. 


These results not only demonstrate the efficacy of our Brain-SubGNN framework in accurately identifying critical regions and structures but also validate its potential as a tool for neuroscientific research. By aligning our findings with existing neurological theories, we establish a credible foundation for the interpretation of our results, offering a promising avenue for further exploration in the domain of brain network analysis.

\section{Discussion}

\subsection{Multicenter experiments with sNC vs. pNC}
\begin{table}[htbp]
  \centering
  \caption{Results of pNC versus sNC classification on ADNI and AIBL, respectively, obtained by the models trained on NACC}
    \begin{tabular}{cccccc}
    \hline\hline
    \textbf{Dataset} & \textbf{pNC/sNC} & \textbf{ACC}   & \textbf{SEN}   & \textbf{SPE}   & \textbf{AUC} \\
    \hline
    ADNI  & 80/249 & 0.675  & 0.588  & 0.703  & 0.644  \\
    AIBL  & 30/59  & 0.730  & 0.600  & 0.797  & 0.739 \\
    \hline\hline
    \end{tabular}%
  \label{tab_multicenter}%
\end{table}%

Given the unexpectedly positive results achieved in the sNC vs. pNC task using a 5-fold cross-validation on the NACC dataset, we extend our approach to additional datasets. We follow the procedures detailed in Section \ref{implementation}. However, both in practice and within the individual data, the sNC group consists of thousands of samples, whereas the pNC group contains significantly fewer samples. As this study does not focus on long-tailed distributions, we select samples within the same interval segment. Specifically, our training set comprises NACC data (281 sNC, 187 pNC), and the independent validation sets includes ADNI-1 (249 sNC, 80 pNC) and AIBL (59 sNC, 11 pNC). 

The results, presented in Table \ref{tab_multicenter}, indicate that the model's performanc on the ADNI and AIBL dataset is considerably inferior to its performance on the NACC dataset. 
We hypothesize that this discrepancy might reflect the inherent challenges of predicting pNC conversion, suggesting that the NACC data may exhibit a high degree of homogeneity. Despite these variations, the significance of this task is evident, and we urge more data collection organizations to prioritize the follow-up data of normal individuals to facilitate the earlier detection of cognitive impairment deterioration.

\subsection{Limitations and future work}
Although the proposed BAGNet has demonstrated promising performance in predicting cognitive impairment conversion, there are still several limitations that need to be addressed in future research.
\textit{First}, our training process is conducted in a segmented manner, requiring the pretraining of the backbone network and clustering fitting in the initial stages. This approach is less efficient compared to an end-to-end framework. Integrating these components into a unified network for simultaneous training could simplify the training process and facilitate complementary guidance between different modalities.
\textit{Second}, the network contains numerous hyperparameters, such as the number of layers and nodes in the brain network, that need to be specified. Our experiments revealed that the optimal number of nodes varies across different stages of cognitive impairment progression. Designing an adaptive method to select the optimal number of nodes would enhance the generalizability and plug-and-play capability of our approach.
\textit{Lastly}, our study has not delved into the clinical interpretability of the detected subgraphs. Future research should focus on providing more detailed interpretations of the identified subgraphs and their implications for understanding the underlying mechanisms of cognitive impairment progression.

\section{Conclusion}
This work presents a framework for constructing dynamic structure brain networks from T1-MRI, emphasizing critical subgraph mining and enhancement with a novel graph representation framework Brain-SubGNN.
The method begins with feature extraction using a convolutional network, followed by constructing the brain network using a correlation matrix between nodes.
Then, Brain-SubGNN adaptively mines and enhances data-specific critical subgraphs, capturing both loop and neighbor subgraphs to reflect long-range and local connections, while maintaining the network's local and global attributes. 
Offering explicit subgraph-level interpretation rather than node- or edge-level interpretation, Brain-SubGNN provides enhanced insights into graph analysis, marking a significant advancement in neuroimaging and cognitive disorder research.

\section*{Acknowledgments}
This work was supported in part by the National Natural Science Foundation of China (Nos.62371499, U23A20483), in part by the Department of Science and Technology of Shandong Province under Grants (SYS202208), and in part by the Guizhou Provincial People's Hospital Talent Fund under Grant Hospital Talent Project ([2022]-5).

Data used in preparation of this article was obtained from the Alzheimers Disease Neuroimaging Initiative (ADNI) database (http://www.adni-info.org/). The investigators within the ADNI contributed to the design and implementation of ADNI and/or provided data, but did not participate in analysis or writing of this report. A complete listing of ADNI investigators can be found at http://adni.loni.usc.edu/wp-content/uploads/how\_to\_apply/ADNI.

\section*{Declaration of Generative AI and AI-assisted technologies in the writing process}
During the preparation of this work the authors used ChatGPT in order to improve readability and language. After using this tool, the authors reviewed and edited the content as needed and take full responsibility for the content of the publication.






\bibliographystyle{elsarticle-num}
\bibliography{refs}

\begin{thebibliography}{10}
\expandafter\ifx\csname url\endcsname\relax
  \def\url#1{\texttt{#1}}\fi
\expandafter\ifx\csname urlprefix\endcsname\relax\def\urlprefix{URL }\fi
\expandafter\ifx\csname href\endcsname\relax
  \def\href#1#2{#2} \def\path#1{#1}\fi

\bibitem{metastasio2006conversion}
A.~Metastasio, P.~Rinaldi, R.~Tarducci, E.~Mariani, F.~T. Feliziani, A.~Cherubini, G.~P. Pelliccioli, G.~Gobbi, U.~Senin, P.~Mecocci, Conversion of mci to dementia: role of proton magnetic resonance spectroscopy, Neurobiology of aging 27~(7) (2006) 926--932.

\bibitem{yue2020characterizing}
L.~Yue, Y.~Pan, T.~Wang, M.~Liu, D.~Shen, S.~Xiao, Characterizing mri biomarkers for early prediction of amnestic mild cognitive impairment among the community-dwelling chinese: Neuroimaging/optimal neuroimaging measures for early detection, Alzheimer's \& Dementia 16 (2020) e041450.

\bibitem{frisoni2010clinical}
G.~B. Frisoni, N.~C. Fox, C.~R. Jack~Jr, P.~Scheltens, P.~M. Thompson, The clinical use of structural mri in alzheimer disease, Nature Reviews Neurology 6~(2) (2010) 67--77.

\bibitem{chincarini2011local}
A.~Chincarini, P.~Bosco, P.~Calvini, G.~Gemme, M.~Esposito, C.~Olivieri, L.~Rei, S.~Squarcia, G.~Rodriguez, R.~Bellotti, et~al., Local mri analysis approach in the diagnosis of early and prodromal alzheimer's disease, Neuroimage 58~(2) (2011) 469--480.

\bibitem{liu2018landmark}
M.~Liu, J.~Zhang, E.~Adeli, D.~Shen, Landmark-based deep multi-instance learning for brain disease diagnosis, Medical image analysis 43 (2018) 157--168.

\bibitem{lian2020attention}
C.~Lian, M.~Liu, Y.~Pan, D.~Shen, Attention-guided hybrid network for dementia diagnosis with structural mr images, IEEE transactions on cybernetics 52~(4) (2020) 1992--2003.

\bibitem{pan2021disease}
Y.~Pan, M.~Liu, Y.~Xia, D.~Shen, Disease-image-specific learning for diagnosis-oriented neuroimage synthesis with incomplete multi-modality data, IEEE transactions on pattern analysis and machine intelligence 44~(10) (2021) 6839--6853.

\bibitem{lian2018hierarchical}
C.~Lian, M.~Liu, J.~Zhang, D.~Shen, Hierarchical fully convolutional network for joint atrophy localization and alzheimer's disease diagnosis using structural mri, IEEE transactions on pattern analysis and machine intelligence 42~(4) (2018) 880--893.

\bibitem{guan2021multi}
H.~Guan, Y.~Liu, E.~Yang, P.-T. Yap, D.~Shen, M.~Liu, Multi-site mri harmonization via attention-guided deep domain adaptation for brain disorder identification, Medical image analysis 71 (2021) 102076.

\bibitem{chen2022alzheimer}
L.~Chen, H.~Qiao, F.~Zhu, Alzheimer's disease diagnosis with brain structural mri using multiview-slice attention and 3d convolution neural network, Frontiers in Aging Neuroscience 14 (2022).

\bibitem{han2023multi}
K.~Han, G.~Li, Z.~Fang, F.~Yang, Multi-template meta-information regularized network for alzheimer’s disease diagnosis using structural mri, IEEE Transactions on Medical Imaging (2023).

\bibitem{zhang2023multi}
J.~Zhang, X.~He, L.~Qing, X.~Chen, Y.~Liu, H.~Chen, Multi-relation graph convolutional network for alzheimer’s disease diagnosis using structural mri, Knowledge-Based Systems 270 (2023) 110546.

\bibitem{zhu2022interpretable}
Y.~Zhu, J.~Ma, C.~Yuan, X.~Zhu, Interpretable learning based dynamic graph convolutional networks for alzheimer’s disease analysis, Information Fusion 77 (2022) 53--61.

\bibitem{bullmore2009complex}
E.~Bullmore, O.~Sporns, Complex brain networks: graph theoretical analysis of structural and functional systems, Nature reviews neuroscience 10~(3) (2009) 186--198.

\bibitem{thiebaut2022emergent}
M.~Thiebaut~de Schotten, S.~J. Forkel, The emergent properties of the connected brain, Science 378~(6619) (2022) 505--510.

\bibitem{axer2022scale}
M.~Axer, K.~Amunts, Scale matters: The nested human connectome, Science 378~(6619) (2022) 500--504.

\bibitem{alvarez2021evolutionary}
U.~Alvarez-Rodriguez, F.~Battiston, G.~F. de~Arruda, Y.~Moreno, M.~Perc, V.~Latora, Evolutionary dynamics of higher-order interactions in social networks, Nature Human Behaviour 5~(5) (2021) 586--595.

\bibitem{chen2024traffic}
J.~Chen, L.~Zheng, Y.~Hu, W.~Wang, H.~Zhang, X.~Hu, Traffic flow matrix-based graph neural network with attention mechanism for traffic flow prediction, Information Fusion 104 (2024) 102146.

\bibitem{fang2022geometry}
X.~Fang, L.~Liu, J.~Lei, D.~He, S.~Zhang, J.~Zhou, F.~Wang, H.~Wu, H.~Wang, Geometry-enhanced molecular representation learning for property prediction, Nature Machine Intelligence 4~(2) (2022) 127--134.

\bibitem{song2021graph}
X.~Song, F.~Zhou, A.~F. Frangi, J.~Cao, X.~Xiao, Y.~Lei, T.~Wang, B.~Lei, Graph convolution network with similarity awareness and adaptive calibration for disease-induced deterioration prediction, Medical Image Analysis 69 (2021) 101947.

\bibitem{song2022multi}
X.~Song, F.~Zhou, A.~F. Frangi, J.~Cao, X.~Xiao, Y.~Lei, T.~Wang, B.~Lei, Multi-center and multi-channel pooling gcn for early ad diagnosis based on dual-modality fused brain network, IEEE Transactions on Medical Imaging (2022).

\bibitem{lei2020self}
B.~Lei, N.~Cheng, A.~F. Frangi, E.-L. Tan, J.~Cao, P.~Yang, A.~Elazab, J.~Du, Y.~Xu, T.~Wang, Self-calibrated brain network estimation and joint non-convex multi-task learning for identification of early alzheimer's disease, Medical image analysis 61 (2020) 101652.

\bibitem{chen2022adversarial}
Y.~Chen, J.~Yan, M.~Jiang, T.~Zhang, Z.~Zhao, W.~Zhao, J.~Zheng, D.~Yao, R.~Zhang, K.~M. Kendrick, et~al., Adversarial learning based node-edge graph attention networks for autism spectrum disorder identification, IEEE Transactions on Neural Networks and Learning Systems (2022).

\bibitem{li2022joint}
Y.~Li, Q.~Wei, E.~Adeli, K.~M. Pohl, Q.~Zhao, Joint graph convolution for analyzing brain structural and functional connectome, in: Medical Image Computing and Computer Assisted Intervention--MICCAI 2022: 25th International Conference, Singapore, September 18--22, 2022, Proceedings, Part I, Springer, 2022, pp. 231--240.

\bibitem{duran2022dual}
F.~S. Duran, A.~Beyaz, I.~Rekik, Dual-hinet: Dual hierarchical integration network of multigraphs for connectional brain template learning, in: Medical Image Computing and Computer Assisted Intervention--MICCAI 2022: 25th International Conference, Singapore, September 18--22, 2022, Proceedings, Part I, Springer, 2022, pp. 305--314.

\bibitem{leng2023dynamic}
Y.~Leng, W.~Cui, C.~Bai, Z.~Chen, Y.~Zheng, J.~Zheng, Dynamic structural brain network construction by hierarchical prototype embedding gcn using t1-mri, in: International Conference on Medical Image Computing and Computer-Assisted Intervention, Springer, 2023, pp. 120--130.

\bibitem{cui2021brainnnexplainer}
H.~Cui, W.~Dai, Y.~Zhu, X.~Li, L.~He, C.~Yang, Brainnnexplainer: An interpretable graph neural network framework for brain network based disease analysis, arXiv preprint arXiv:2107.05097 (2021).

\bibitem{kan2022fbnetgen}
X.~Kan, H.~Cui, J.~Lukemire, Y.~Guo, C.~Yang, Fbnetgen: Task-aware gnn-based fmri analysis via functional brain network generation, in: International Conference on Medical Imaging with Deep Learning, PMLR, 2022, pp. 618--637.

\bibitem{zhu2022joint}
Y.~Zhu, H.~Cui, L.~He, L.~Sun, C.~Yang, Joint embedding of structural and functional brain networks with graph neural networks for mental illness diagnosis, in: 2022 44th Annual International Conference of the IEEE Engineering in Medicine \& Biology Society (EMBC), IEEE, 2022, pp. 272--276.

\bibitem{cui2022interpretable}
H.~Cui, W.~Dai, Y.~Zhu, X.~Li, L.~He, C.~Yang, Interpretable graph neural networks for connectome-based brain disorder analysis, in: International Conference on Medical Image Computing and Computer-Assisted Intervention, Springer, 2022, pp. 375--385.

\bibitem{kriege2020survey}
N.~M. Kriege, F.~D. Johansson, C.~Morris, A survey on graph kernels, Applied Network Science 5~(1) (2020) 1--42.

\bibitem{shervashidze2011weisfeiler}
N.~Shervashidze, P.~Schweitzer, E.~J. Van~Leeuwen, K.~Mehlhorn, K.~M. Borgwardt, Weisfeiler-lehman graph kernels., Journal of Machine Learning Research 12~(9) (2011).

\bibitem{chen2022structure}
D.~Chen, L.~O’Bray, K.~Borgwardt, Structure-aware transformer for graph representation learning, in: International Conference on Machine Learning, PMLR, 2022, pp. 3469--3489.

\bibitem{geisler2023transformers}
S.~Geisler, Y.~Li, D.~J. Mankowitz, A.~T. Cemgil, S.~G{\"u}nnemann, C.~Paduraru, Transformers meet directed graphs, in: International Conference on Machine Learning, PMLR, 2023, pp. 11144--11172.

\bibitem{sitaram2017closed}
R.~Sitaram, T.~Ros, L.~Stoeckel, S.~Haller, F.~Scharnowski, J.~Lewis-Peacock, N.~Weiskopf, M.~L. Blefari, M.~Rana, E.~Oblak, et~al., Closed-loop brain training: the science of neurofeedback, Nature Reviews Neuroscience 18~(2) (2017) 86--100.

\bibitem{skouras2020earliest}
S.~Skouras, J.~Torner, P.~Andersson, Y.~Koush, C.~Falcon, C.~Minguillon, K.~Fauria, F.~Alpiste, K.~Blenow, H.~Zetterberg, et~al., Earliest amyloid and tau deposition modulate the influence of limbic networks during closed-loop hippocampal downregulation, Brain 143~(3) (2020) 976--992.

\bibitem{van2011rich}
M.~P. Van Den~Heuvel, O.~Sporns, Rich-club organization of the human connectome, Journal of Neuroscience 31~(44) (2011) 15775--15786.

\bibitem{bassett2006small}
D.~S. Bassett, E.~Bullmore, Small-world brain networks, The neuroscientist 12~(6) (2006) 512--523.

\bibitem{ashburner2000voxel}
J.~Ashburner, K.~J. Friston, Voxel-based morphometry—the methods, Neuroimage 11~(6) (2000) 805--821.

\bibitem{zhang2011multimodal}
D.~Zhang, Y.~Wang, L.~Zhou, H.~Yuan, D.~Shen, A.~D.~N. Initiative, et~al., Multimodal classification of alzheimer's disease and mild cognitive impairment, Neuroimage 55~(3) (2011) 856--867.

\bibitem{bronstein2017geometric}
M.~M. Bronstein, J.~Bruna, Y.~LeCun, A.~Szlam, P.~Vandergheynst, Geometric deep learning: going beyond euclidean data, IEEE Signal Processing Magazine 34~(4) (2017) 18--42.

\bibitem{gartner2003graph}
T.~G{\"a}rtner, P.~Flach, S.~Wrobel, On graph kernels: Hardness results and efficient alternatives, in: Learning Theory and Kernel Machines: 16th Annual Conference on Learning Theory and 7th Kernel Workshop, COLT/Kernel 2003, Washington, DC, USA, August 24-27, 2003. Proceedings, Springer, 2003, pp. 129--143.

\bibitem{wang2021inductive}
Y.~Wang, Y.-Y. Chang, Y.~Liu, J.~Leskovec, P.~Li, Inductive representation learning in temporal networks via causal anonymous walks, International Conference on Learning Representations (2021).

\bibitem{borgwardt2005shortest}
K.~M. Borgwardt, H.-P. Kriegel, Shortest-path kernels on graphs, in: Fifth IEEE international conference on data mining (ICDM'05), IEEE, 2005, pp. 8--pp.

\bibitem{luo2022neighborhood}
Y.~Luo, P.~Li, Neighborhood-aware scalable temporal network representation learning, in: Learning on Graphs Conference, PMLR, 2022, pp. 1--1.

\bibitem{cong2023we}
W.~Cong, S.~Zhang, J.~Kang, B.~Yuan, H.~Wu, X.~Zhou, H.~Tong, M.~Mahdavi, Do we really need complicated model architectures for temporal networks?, arXiv preprint arXiv:2302.11636 (2023).

\bibitem{tang2023systematic}
Z.~Tang, T.~Li, D.~Wu, J.~Liu, Z.~Yang, A systematic literature review of reinforcement learning-based knowledge graph research, Expert Systems with Applications (2023) 121880.

\bibitem{lai2020policy}
K.-H. Lai, D.~Zha, K.~Zhou, X.~Hu, Policy-gnn: Aggregation optimization for graph neural networks, in: Proceedings of the 26th ACM SIGKDD International Conference on Knowledge Discovery \& Data Mining, 2020, pp. 461--471.

\bibitem{xu2020adversarial}
H.~Xu, Y.~Ma, H.-C. Liu, D.~Deb, H.~Liu, J.-L. Tang, A.~K. Jain, Adversarial attacks and defenses in images, graphs and text: A review, International Journal of Automation and Computing 17 (2020) 151--178.

\bibitem{zhu2019causal}
S.~Zhu, I.~Ng, Z.~Chen, Causal discovery with reinforcement learning, arXiv preprint arXiv:1906.04477 (2019).

\bibitem{yuan2021explainability}
H.~Yuan, H.~Yu, J.~Wang, K.~Li, S.~Ji, On explainability of graph neural networks via subgraph explorations, in: International conference on machine learning, PMLR, 2021, pp. 12241--12252.

\bibitem{mazyavkina2021reinforcement}
N.~Mazyavkina, S.~Sviridov, S.~Ivanov, E.~Burnaev, Reinforcement learning for combinatorial optimization: A survey, Computers \& Operations Research 134 (2021) 105400.

\bibitem{wang2020second}
Z.~Wang, S.~Ji, Second-order pooling for graph neural networks, IEEE Transactions on Pattern Analysis and Machine Intelligence (2020).

\bibitem{dou2020enhancing}
Y.~Dou, Z.~Liu, L.~Sun, Y.~Deng, H.~Peng, P.~S. Yu, Enhancing graph neural network-based fraud detectors against camouflaged fraudsters, in: Proceedings of the 29th ACM international conference on information \& knowledge management, 2020, pp. 315--324.

\bibitem{peng2022reinforced}
H.~Peng, R.~Zhang, S.~Li, Y.~Cao, S.~Pan, S.~Y. Philip, Reinforced, incremental and cross-lingual event detection from social messages, IEEE Transactions on Pattern Analysis and Machine Intelligence 45~(1) (2022) 980--998.

\bibitem{mnih2015human}
V.~Mnih, K.~Kavukcuoglu, D.~Silver, A.~A. Rusu, J.~Veness, M.~G. Bellemare, A.~Graves, M.~Riedmiller, A.~K. Fidjeland, G.~Ostrovski, et~al., Human-level control through deep reinforcement learning, nature 518~(7540) (2015) 529--533.

\bibitem{gao2019graphnas}
Y.~Gao, H.~Yang, P.~Zhang, C.~Zhou, Y.~Hu, Graphnas: Graph neural architecture search with reinforcement learning, arXiv preprint arXiv:1904.09981 (2019).

\bibitem{zheng2017learning}
H.~Zheng, J.~Fu, T.~Mei, J.~Luo, Learning multi-attention convolutional neural network for fine-grained image recognition, in: Proceedings of the IEEE international conference on computer vision, 2017, pp. 5209--5217.

\bibitem{lloyd1982least}
S.~Lloyd, Least squares quantization in pcm, IEEE transactions on information theory 28~(2) (1982) 129--137.

\bibitem{li2020prototypical}
J.~Li, P.~Zhou, C.~Xiong, S.~C. Hoi, Prototypical contrastive learning of unsupervised representations, arXiv preprint arXiv:2005.04966 (2020).

\bibitem{vaswani2017attention}
A.~Vaswani, N.~Shazeer, N.~Parmar, J.~Uszkoreit, L.~Jones, A.~N. Gomez, {\L}.~Kaiser, I.~Polosukhin, Attention is all you need, Advances in neural information processing systems 30 (2017).

\bibitem{petersen2010alzheimer}
R.~C. Petersen, P.~S. Aisen, L.~A. Beckett, M.~C. Donohue, A.~C. Gamst, D.~J. Harvey, C.~R. Jack, W.~J. Jagust, L.~M. Shaw, A.~W. Toga, et~al., Alzheimer's disease neuroimaging initiative (adni): clinical characterization, Neurology 74~(3) (2010) 201--209.

\bibitem{besser2018version}
L.~Besser, W.~Kukull, D.~S. Knopman, H.~Chui, D.~Galasko, S.~Weintraub, G.~Jicha, C.~Carlsson, J.~Burns, J.~Quinn, et~al., Version 3 of the national alzheimer’s coordinating center’s uniform data set, Alzheimer Disease \& Associated Disorders 32~(4) (2018) 351--358.

\bibitem{trockman2022patches}
A.~Trockman, J.~Z. Kolter, Patches are all you need?, arXiv preprint arXiv:2201.09792 (2022).

\bibitem{chen2020improved}
X.~Chen, H.~Fan, R.~Girshick, K.~He, Improved baselines with momentum contrastive learning, arXiv preprint arXiv:2003.04297 (2020).

\bibitem{li2021deepgcns}
G.~Li, M.~M{\"u}ller, G.~Qian, I.~C.~D. Perez, A.~Abualshour, A.~K. Thabet, B.~Ghanem, Deepgcns: Making gcns go as deep as cnns, IEEE transactions on pattern analysis and machine intelligence (2021).

\bibitem{xu2018powerful}
K.~Xu, W.~Hu, J.~Leskovec, S.~Jegelka, How powerful are graph neural networks?, arXiv preprint arXiv:1810.00826 (2018).

\bibitem{du2001magnetic}
A.~Du, N.~Schuff, D.~Amend, M.~Laakso, Y.~Hsu, W.~Jagust, K.~Yaffe, J.~Kramer, B.~Reed, D.~Norman, et~al., Magnetic resonance imaging of the entorhinal cortex and hippocampus in mild cognitive impairment and alzheimer's disease, Journal of Neurology, Neurosurgery \& Psychiatry 71~(4) (2001) 441--447.

\bibitem{pini2016brain}
L.~Pini, M.~Pievani, M.~Bocchetta, D.~Altomare, P.~Bosco, E.~Cavedo, S.~Galluzzi, M.~Marizzoni, G.~B. Frisoni, Brain atrophy in alzheimer’s disease and aging, Ageing research reviews 30 (2016) 25--48.

\bibitem{hampstead2023toward}
B.~M. Hampstead, A.~Y. Stringer, A.~D. Iordan, R.~Ploutz-Snyder, K.~Sathian, Toward rational use of cognitive training in those with mild cognitive impairment, Alzheimer's \& Dementia 19~(3) (2023) 933--945.

\end{thebibliography}

\end{document}